%% file: main.tex
\documentclass[letterpaper]{article} 
\makeatletter
\def\@fnsymbol#1{\ensuremath{\ifcase#1\or \dagger\or \ddagger\or
   \mathsection\or \mathparagraph\or \|\or **\or \dagger\dagger
   \or \ddagger\ddagger \else\@ctrerr\fi}}
\makeatother
\usepackage{aaai2026}  
\usepackage{times}  
\usepackage{helvet}  
\usepackage{courier}  
\usepackage[hyphens]{url}  
\usepackage{graphicx} 
\usepackage{natbib}  
\usepackage{caption} 
\usepackage{algorithm}
\usepackage{algorithmic}

\usepackage{newfloat}
\usepackage{listings}
\DeclareCaptionStyle{ruled}{labelfont=normalfont,labelsep=colon,strut=off} 
\floatstyle{ruled}
\newfloat{listing}{tb}{lst}{}
\floatname{listing}{Listing}
\title{DICE: Distilling Classifier-Free Guidance into Text Embeddings}
\author {
    Zhenyu Zhou\textsuperscript{\rm 1,\rm 2},
    Defang Chen\textsuperscript{\rm 3}\thanks{Correspondence to: Defang Chen \textless defangch@buffalo.edu\textgreater},
    Can Wang\textsuperscript{\rm 1,\rm 2},
    Chun Chen\textsuperscript{\rm 1,\rm 2},
    Siwei Lyu\textsuperscript{\rm 3}
}
\affiliations {
    \textsuperscript{\rm 1}Zhejiang University, State Key Laboratory of Blockchain and Data Security\\
    \textsuperscript{\rm 2}Hangzhou High-Tech Zone (Binjiang) Institute of Blockchain and Data Security\\
    \textsuperscript{\rm 3}University at Buffalo, State University of New York
}

\usepackage{bibentry}

\usepackage{amssymb}
\usepackage{amsmath}
\usepackage{xspace}
\usepackage{booktabs}
\usepackage{subcaption}
\usepackage{multirow}
\usepackage{colortbl}
\usepackage{xcolor}

\def\bfc{\mathbf{c}}

\def\bff{\mathbf{f}}

\def\bfs{\mathbf{s}}

\def\bfw{\mathbf{w}}
\def\bfx{\mathbf{x}}

\def\bfI{\mathbf{I}}

\def\bfeps{\boldsymbol{\epsilon}}

\def\rmd{\mathrm{d}}
\def\rmdx{\mathrm{d}\bfx}
\def\bbE{\mathbb{E}}
\def\bbR{\mathbb{R}}

\def\eqref#1{Eq.~(\ref{#1})}

\def\ourName{DICE\xspace}
\begin{document}

\maketitle

\maketitle

\input{secs/abstract}
\input{secs/intro}
\input{secs/background}

\input{secs/method}
\input{secs/exp}
\input{secs/related}
\input{secs/conclusion}

\section{Acknowledgments}
Zhenyu Zhou and Can Wang are supported by the National Natural Science Foundation of China (No. 62476244), the Starry Night Science Fund of Zhejiang University Shanghai Institute for Advanced Study, China (Grant No: SN-ZJU-SIAS-001) and the advanced computing resources provided by the Supercomputing Center of Hangzhou City University.

\bibliography{aaai2026}
\input{secs/appendix}

\end{document}

%% file: secs/abstract.tex
\begin{abstract}

Text-to-image diffusion models are capable of generating high-quality images, but suboptimal pre-trained text representations often result in these images failing to align closely with the given text prompts. Classifier-free guidance (CFG) is a popular and effective technique for improving text-image alignment in the generative process. However, CFG introduces significant computational overhead.
In this paper, we present \textbf{D}\textbf{I}stilling \textbf{C}FG by sharpening text \textbf{E}mbeddings (\ourName) that replaces CFG in the sampling process with half the computational complexity while maintaining similar generation quality. 
\ourName distills a CFG-based text-to-image diffusion model into a CFG-free version by refining text embeddings to replicate CFG-based directions. 
In this way, we avoid the computational drawbacks of CFG, enabling high-quality, well-aligned image generation at a fast sampling speed.
Furthermore, examining the enhancement pattern, we identify the underlying mechanism of \ourName that sharpens specific components of text embeddings to preserve semantic information while enhancing fine-grained details. 
Extensive experiments on multiple Stable Diffusion v1.5 variants, SDXL, and PixArt-$\alpha$ demonstrate the effectiveness of our method. 
Code is available at \url{https://github.com/zju-pi/dice}.
\end{abstract}


%% file: secs/intro.tex
\section{Introduction}
\label{intro}

Diffusion-based generative models~\cite{sohl2015deep,song2019ncsn,ho2020ddpm} have recently achieved remarkable advances, driven by continuously refined theoretical frameworks~\cite{song2021sde,karras2022edm,chen2024trajectory,kingma2024understanding} and the fast evolution of model architectures~\cite{peebles2023scalable,bao2023all}.
Their impressive generation ability brings text-to-image generation to unprecedented levels~\cite{rombach2022ldm,saharia2022photorealistic,podell2024sdxl,esser2024scaling}, and enables a multitude of new conditional generation tasks~\cite{croitoru2023diffusion}.

\begin{figure}[t!]
    \centering
    \includegraphics[width=0.98\columnwidth]{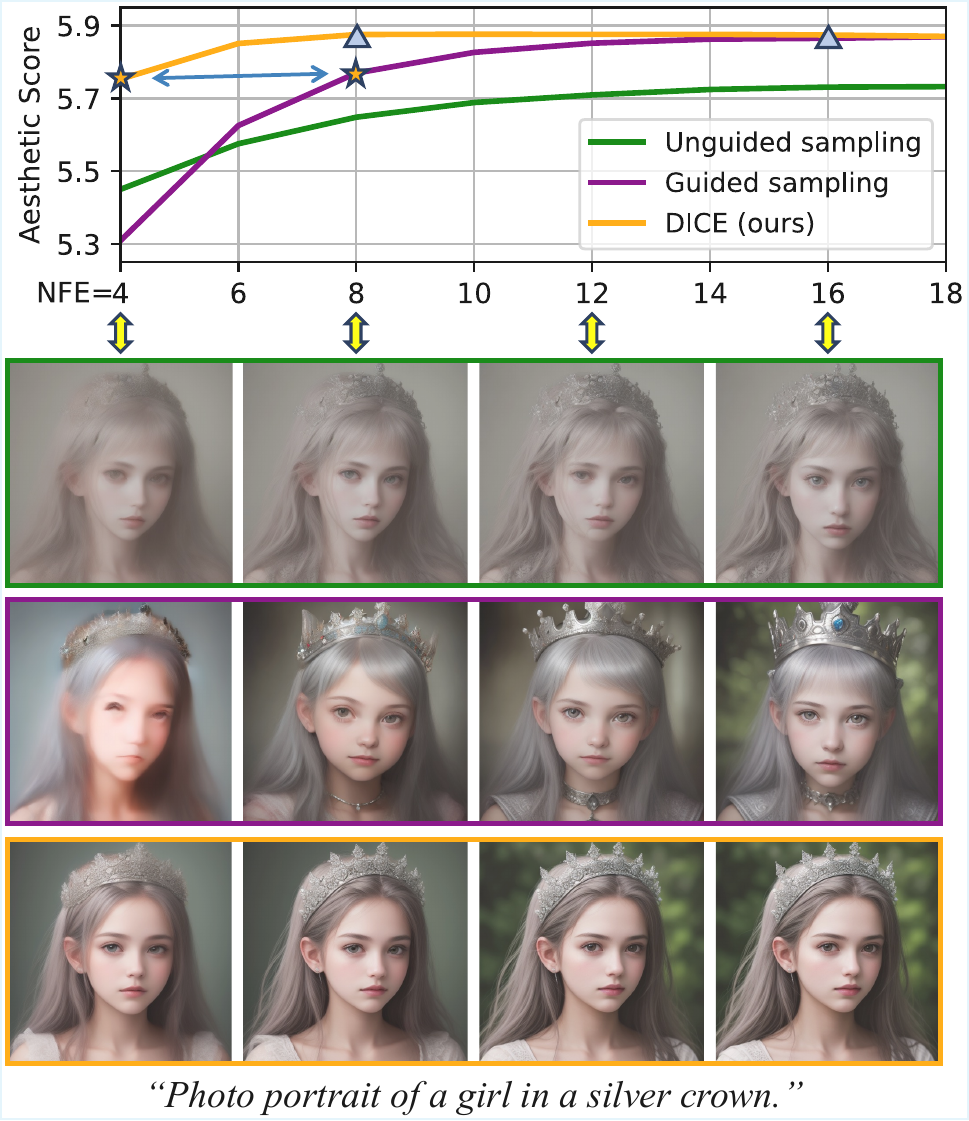}
    \caption{Comparison of text-to-image generation: unguided sampling, guided sampling, and DICE.  \textit{Top}: Average aesthetic score~\cite{schuhmann2022laion} over $5,000$ images plotted against the number of function evaluations (NFE). \textit{Bottom}: An example of image synthesis using different methods at NFE = 4, 8, 12, and 16. 
    }
    \label{fig:teaser}
\end{figure}

In text-to-image generation~\cite{nichol2022glide,rombach2022ldm,saharia2022photorealistic}, diffusion models use text embeddings produced by pre-trained encoders such as CLIP~\cite{radford2021learning} and T5~\cite{raffel2020exploring}. These embeddings are fixed-dimensional vectors that encapsulate the semantic content of text prompts. However, they are not specifically optimized for image generation~\cite{li2024textcraftor}. Moreover, images often encompass more detailed information than text prompts can convey, making precise text-image semantic alignment challenging~\cite{schrodi2024two}. Consequently, as illustrated in Figure \ref{fig:teaser}, sampling with text-to-image models in their original conditional form—hereafter referred to as \textit{unguided sampling}—often produces blurry and semantically inaccurate outputs~\cite{meng2023distillation,karras2024guiding}. To address the limited semantic signals provided by text embeddings, \textit{guided sampling} techniques~\cite{dhariwal2021diffusion,ho2022classifier} have been introduced to steer samples toward a more concentrated distribution. 
{\it Classifier-Free Guidance} (CFG)~\cite{ho2022classifier} is a widely adopted technique for guided sampling. It directs the generative process at each sampling step by extrapolating the direction between the conditional prediction and an unconditional prediction, with the guidance strength modulated by a hyperparameter known as the guidance scale. CFG enhances both image quality and text-image alignment, making it a popular choice in practice. However, an important drawback of CFG is that it requires an additional model evaluation at each step, thereby increasing the sampling overhead~\cite{ho2022classifier}. 
Moreover, since CFG deviates from the sampling path of a normal diffusion model, it complicates the understanding of sampling dynamics~\cite{karras2024guiding,zheng2024characteristic,bradley2024classifier}. 

To mitigate the increased sampling overhead, prior research distilled CFG into a single model evaluation per sampling step~\cite{meng2023distillation,hsiao2024plug}. 
While these methods can effectively reduce the computational cost of CFG, they typically incur significant training overhead due to the large number of trainable parameters required and suffer from practical issues.
For instance, on the Stable Diffusion v1.5 model~\cite{rombach2022ldm}, Guided Distillation (GD)~\cite{meng2023distillation} fine-tunes the whole model involving 859M trainable parameters, and the fine-tuned model cannot be applied to new scenarios.
Plug-and-Play Distillation (PnP)~\cite{hsiao2024plug} trains an auxiliary model with 361M parameters but requires multiple operations during inference, reducing the ratio of acceleration.

\begin{figure}[t]
    \centering
    \includegraphics[width=\columnwidth]{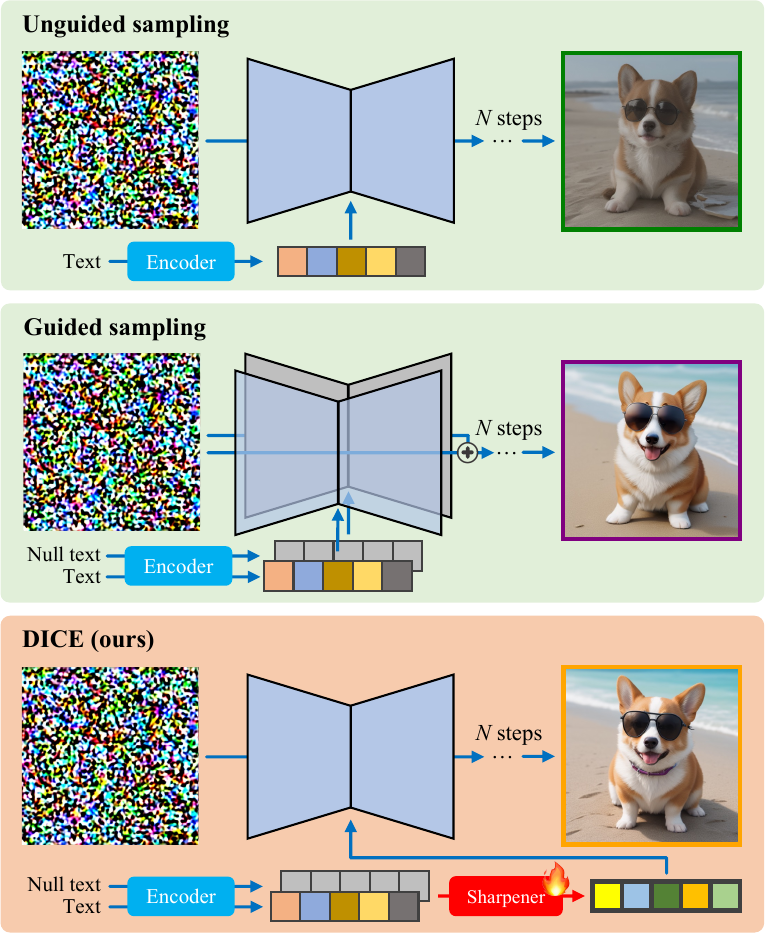}
    \caption{Overview of DICE sampling and comparison with traditional unguided and guided sampling. With sharpened text embeddings, DICE achieves high-quality image generation comparable to guided sampling while maintaining the same computational overhead as unguided sampling.}
    \label{fig:comparison}
\end{figure}

In this paper, we introduce \textbf{D}\textbf{I}stilling \textbf{C}FG by sharpening text \textbf{E}mbeddings (\ourName) as an alternative approach for achieving high-quality image generation with unguided sampling. Specifically, we refine the model's input condition, \textit{i.e.}, text embeddings, under CFG-based supervision by training a lightweight sharpener that operates only once independently of the primary text-to-image model with only 2M model parameters (Figure \ref{fig:comparison}). With sharpened embeddings, our enhanced unguided sampling achieves image quality on par with guided sampling while maintaining computational efficiency. 
By inspecting the underlying mechanism, we reveal that \ourName identifies a universal enhancement pattern: the semantically irrelevant components of the text embedding are primarily amplified, preserving essential semantic information while enriching fine-grained details in the generated images. Extensive experiments across various text-to-image models, encompassing different model capacities, image styles, and network architectures, validate the effectiveness of our method in diverse scenarios. 

%% file: secs/background.tex
\section{Preliminaries}

\begin{figure*}[t]
    \centerline{\includegraphics[width=\textwidth]{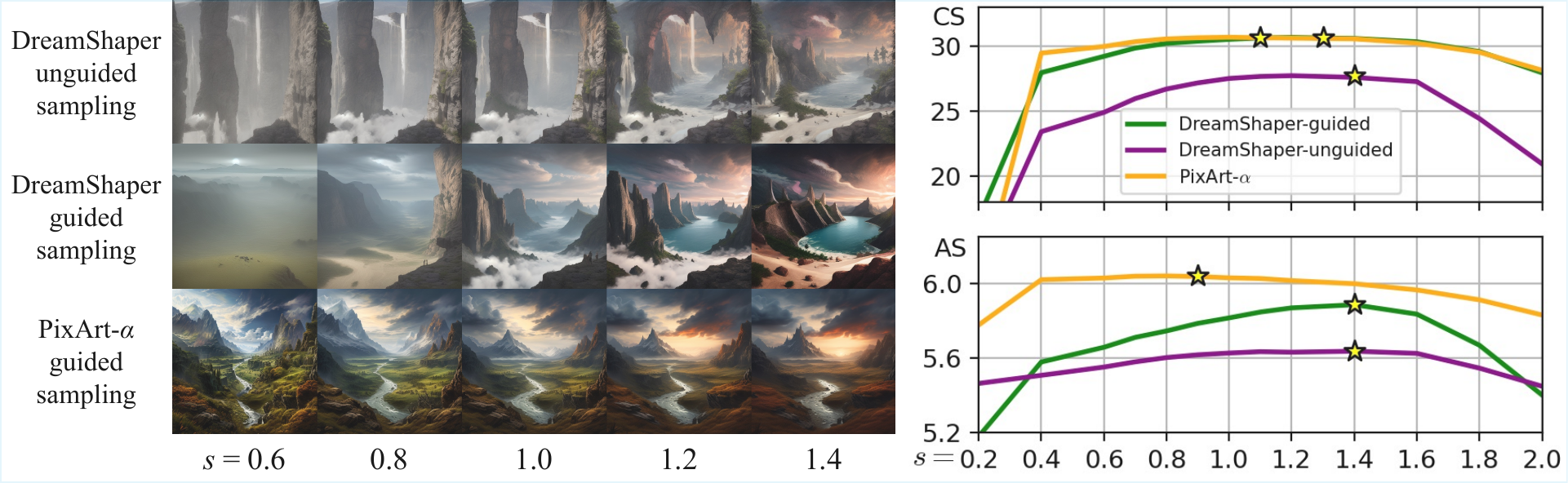}}
    \caption{Text-image alignment with scaled text embeddings. Images are generated by DreamShaper~\cite{DreamShaper}, a popular variant of Stable Diffusion v1.5~\cite{rombach2022ldm} with a CLIP text encoder~\cite{radford2021learning}, and PixArt-$\alpha$~\cite{chen2024pixart} with a T5-XXL text encoder~\cite{raffel2020exploring}. 
    \textit{Left}: Text embeddings are scaled by a factor $s$ and images are generated via unguided and guided sampling. 
    \textit{Right}: A grid search is conducted to identify the optimal scaling factor with respect to the CLIP score (CS) and Aesthetic score (AS). 
    An optimal scaling factor improves the sample quality but varies across models. Meanwhile, naive scaling is insufficient to improve unguided sampling to the image quality achieved by guided sampling, which necessitates exploring the embedding space for a fine-grained dynamic scaling.
    Prompt: ``\textit{An epic landscape}''.
    }
    \label{fig:motivation}
\end{figure*}

\subsection{Diffusion Models}
Given a data sample $\bfx_0 \in \mathbb{R}^d$ from the implicit target data distribution $p_0$ (in this case, the distribution of all natural images), the forward process in diffusion models gradually adds white Gaussian noise to the sample, following a stochastic differential equation (SDE)~\cite{song2021sde}: $\rmdx_t = \bff(\bfx_t, t)\rmd t + g(t) \rmd \bfw_t$,
where $t \in \left[0, T\right]$, $\bff(\cdot, t): \bbR^{d} \rightarrow \bbR^d, g(\cdot): \bbR \rightarrow \bbR$ are drift and diffusion coefficients and $\bfw_t \in \bbR^d$ is the Wiener process~\cite{oksendal2013stochastic}. 
The backward process in diffusion models achieves the data reconstruction through a reverse-time SDE, $\rmdx_t = [\bff(\bfx_t, t)- g^2(t)\nabla_{\bfx_t} \log p_{t}(\bfx_t)]\rmd t + g(t) \rmd \bar{\bfw}_t$, which shares the same marginal distributions $\{p_t\}_{t=0}^T$ with the forward process~\cite{song2021sde}. This reverse-time SDE has a \textit{probability flow} ordinary differential equation (PF-ODE) counterpart~\cite{song2021sde,chen2024trajectory}, $\rmdx_t = [\bff(\bfx_t, t)- \frac{1}{2}g^2(t)\nabla_{\bfx_t} \log p_{t}(\bfx_t)]\rmd t$. Following the parametrization in EDM~\cite{karras2022edm}, where $\bff(\bfx_t, t)=\mathbf{0}$ and $g(t)=\sqrt{2t}$, we simplify the PF-ODE into
\begin{equation}
    \label{eq:pf_ode}
    \rmdx_t = -t \nabla_{\bfx_t} \log p_{t}(\bfx_t) \rmd t.
\end{equation}
The analytically intractable $\nabla_{\bfx_t} \log p_{t}(\bfx_t)$ is known as the \textit{score function}~\cite{hyvarinen2005estimation,lyu2009interpretation}, which is typically estimated by either a score-prediction model $\bfs_\theta(\bfx_t)$, or a noise-prediction model $\bfeps_{\theta}(\bfx_t)$, i.e., 
\begin{equation}
    \label{eq:equalvalent}
    \nabla_{\bfx_t} \log p_{t}(\bfx_t) \approx \bfs_\theta(\bfx_t) = -\frac{\bfeps_{\theta}(\bfx_t)}{t}.
\end{equation}
For simplicity, unless otherwise specified, we will drop the time dependence of the model subsequently to reduce notational clutter. 
The training objective of diffusion models is a weighted minimization of a regression loss~\cite{ho2020ddpm,nichol2021improved,kingma2024understanding}. For distillation tasks in which a student model $\bfeps_{\theta}$ is supervised by a fixed teacher model $\tilde{\bfeps}_{\tilde{\theta}}$, the training objective is defined as $\mathcal{L}(\theta) = \mathbb{E}_{t \sim \mathcal{U}(0,T), \bfeps \sim \mathcal{N}(\mathbf{0},\bfI)} \left[\lambda(t) \lVert \bfeps_{\theta}(\bfx_t) - \tilde{\bfeps}_{\tilde{\theta}}(\bfx_t) \rVert\right]$,
where $\lambda(t)$ is a weighting function, $\bfx_t = \bfx_0 + t \bfeps$, and $\bfx_0 \sim p_0$ follows the forward transition kernel $p_{0t}\left(\bfx_t|\bfx_0\right) = \mathcal{N}\left(\bfx_t;\bfx_0,t^2\bfI\right)$.

In text-to-image generation, the diffusion model receives embeddings of a text prompt $\bfc \in \bbR^{K\times d_e}$ encoded by a pre-trained text encoder to predict the score function conditioned on the text prompt, where $K$ denotes the token number and $d_e$ is the context dimension of each token. Starting from a random Gaussian noise $\bfx_T$ with a manually designed time schedule, sampling from diffusion models is to numerically solve $\rmdx_t = \bfeps_{\theta}(\bfx_t,\bfc) \rmd t$ through, for example, an Euler discretization~\cite{song2021ddim},
\begin{equation}
    \label{eq:euler}
    \bfx_s = \bfx_t + \left(s-t\right)\bfeps_{\theta}(\bfx_t,\bfc),
\end{equation}
where $0 \leq s < t \leq T$. Advanced numerical solvers using higher-order derivatives can also be employed to achieve accelerated sampling of diffusion models~\cite{zhang2023deis,zhou2023fast}.

\subsection{Classifier-free Guidance}
The standard class-conditional sampling for text-to-image generation with Equation \ref{eq:euler} usually produces blurry, distorted, and semantically inaccurate images~\cite{meng2023distillation,karras2024guiding}.
In practice, classifier-free guidance (CFG)~\cite{ho2022classifier} is widely used to trade sample fidelity for diversity, allowing the model to achieve low-temperature sampling without the need for an auxiliary classifier-based guidance~\cite{dhariwal2021diffusion}. This technique modifies the model output by another model evaluation conditioned on a fixed null text embedding $\bfc_\text{null}$:
\begin{equation}
    \label{eq:cfg}
    \bfeps_{\theta}^{\omega,\bfc_\text{null}}(\bfx_t,\bfc) = \omega \bfeps_{\theta}(\bfx_t,\bfc) - \left(\omega - 1\right)  \bfeps_{\theta}(\bfx_t,\bfc_\text{null}),
\end{equation}
\begin{equation}
    \label{eq:euler_cfg}
    \bfx_s = \bfx_t + \left(s-t\right)\bfeps_{\theta}^{\omega,\bfc_\text{null}}(\bfx_t,\bfc),
\end{equation}
where $\omega \geq 1$ is known as the \textit{guidance scale}, with $\omega=1$ corresponding to unguided sampling, and $\omega > 1$ to guided sampling. Despite the ability to perform high-quality generation, CFG requires one more model evaluation in each guided sampling step, increasing the inference costs. 



%% file: secs/method.tex
\section{Method}
\label{sec:method}

\subsection{Sharpening Text Embeddings by Scaling}
\label{sec:motivation}

Text-to-image diffusion models are trained on large datasets of text-image pairs~\cite{rombach2022ldm,podell2024sdxl,esser2024scaling}. In this process, text prompts are first encoded into embeddings using pre-trained text encoders~\cite{radford2021learning,raffel2020exploring} and then integrated into the model inference via cross-attention modules. However, these models often struggle to generate images that closely align with the input prompts when using unguided sampling.

We hypothesize that this misalignment stems from two primary factors. 
First, current text encoders are not specifically designed for image generation. CLIP models align text and images in the embedding space via contrastive learning~\cite{radford2021learning}, while T5 models are fine-tuned on large-scale natural language processing tasks~\cite{raffel2020exploring}. Neither is optimized to provide text embeddings tailored for high-quality image generation.
Second, there is an inherent information imbalance between text and images. Images encapsulate rich details such as layout, texture, and fine-grained elements, whereas manually annotated captions typically describe only the main concepts~\cite{radford2021learning,schuhmann2022laion}. This disparity leads to a well-known modality gap between text and image domains~\cite{liang2022mind,schrodi2024two}, particularly when the text prompt length is limited. This may result in subpar sample quality in unguided sampling.


Instead of relying on CFG in the sampling process with double computational overhead, we improve the text-image alignment by sharpening the text embeddings. 
We begin by verifying the existence of such embeddings using the most straightforward approach: scaling.
In Figure \ref{fig:motivation}, we scale the text embeddings $\bfc$ input to the text-to-image models by a factor $s$ and apply the scaled embeddings to both unguided (Equation \ref{eq:euler}) and guided (Equation \ref{eq:euler_cfg}) sampling. Sharpened text embeddings yield enriched image details and improved image contrast, but the optimal scaling factor varies across text-to-image models and text prompts. As illustrated in Figure \ref{fig:motivation}, scaling factors of 0.6 and 1.4 can both enhance image details. 
Naive scaling alone is insufficient for improving unguided sampling to the level of image quality achieved by guided sampling. 
However, our pilot experiment demonstrates that while text-to-image models are trained on pre-trained text embeddings, they can generalize to a broader embedding space, making optimal sharpened text embeddings worth exploring. 
To learn the patterns of sharpened text embeddings that can more effectively improve text-image alignment, we propose training a lightweight neural network to dynamically scale the text embeddings.


\begin{algorithm}[t]
\caption{\ourName Training}
\label{alg:alg}
\begin{algorithmic}
    \STATE {\bfseries Input:} dataset $\mathcal{D}$, guidance scale $\omega$, maximum timestamp $T$, text-to-image model $\bfeps_{\theta}(\cdot,\cdot)$, null text embedding $\bfc_\text{null}$, learning rate $\eta$
    \STATE {\bfseries Initialize:} sharpener $r_{\phi}(\cdot,\cdot)$
    \WHILE{not converged}
    \STATE Sample image-embedding pairs $(\bfx_0, \bfc) \sim D$
    \STATE Sample a timestamp $t \sim \mathcal{U}(0, T)$
    \STATE Forward diffusion process $\bfx_t \sim \mathcal{N}(\bfx_0, t^2I)$
    \STATE $\bfeps_{\theta}^{\omega,\bfc_{\text{null}}}(\bfx_t,\bfc) = \omega \bfeps_{\theta}(\bfx_t,\bfc) - \left(\omega - 1\right)  \bfeps_{\theta}(\bfx_t,\bfc_{\text{null}})$
    \STATE $\bfc_\phi = \bfc + \alpha r_{\phi}(\bfc,\bfc_\text{null})$
    \STATE $\mathcal{L}(\phi) = \lVert \bfeps_{\theta}(\bfx_t,\bfc_\phi) - \bfeps_{\theta}^{\omega,\bfc_\text{null}}(\bfx_t,\bfc) \rVert$
    \STATE $\phi \leftarrow \phi - \eta \nabla_{\phi}\mathcal{L}(\phi)$
    \ENDWHILE
\end{algorithmic}
\end{algorithm}

\subsection{\ourName}
\label{sec:dice}

We present \textbf{D}\textbf{I}stilling \textbf{C}FG by sharpening text \textbf{E}mbeddings (\ourName) which enhances unguided sampling by aligning its sampling trajectory with the CFG trajectory.
As such, \ourName cuts the computational cost of CFG in half as it calls the denoising model only once per sampling step, while keeping the high generation quality of CFG.
Specifically, given a text embedding $\bfc$ encoded by the text encoder, we train a lightweight sharpener $r_{\phi}(\cdot,\cdot): \bbR^{(K\times d_e) \times (K\times d_e)} \rightarrow \bbR^{K \times d_e}$ with the trainable parameters $\phi$, to sharpen the original text embedding, i.e., 
\begin{equation}
    \label{eq:sharpener}
    \bfc_{\phi} = \bfc + \alpha r_{\phi}(\bfc,\bfc_\text{null}),
\end{equation}
where $\alpha$ is a hyperparameter controlling the sharpening strength. Similar to Equation \ref{eq:euler}, the unguided sampling becomes $\bfx_s = \bfx_t + \left(s-t\right)\bfeps_{\theta}(\bfx_t,\bfc_{\phi})$.
We obtain the sharpened text embedding using CFG-based supervision while keeping the original text-to-image model frozen. 
Given image-embedding pairs $(\bfx_0, \bfc)$, the training loss for the sharpener is formulated in a distillation manner as:
\begin{equation}
    \label{eq:loss}
    \bbE_{t \sim \mathcal{U}(0,T), \bfx_t \sim \mathcal{N}(\bfx_0,t^2\bfI)}\lVert \bfeps_{\theta}(\bfx_t,\bfc_\phi) - \bfeps_{\theta}^{\omega,\bfc_\text{null}}(\bfx_t,\bfc) \rVert,
\end{equation}
where the trainable parameter is $\phi$, and $\theta$ remains fixed. 
The training procedure is described in Algorithm \ref{alg:alg}.
As shown in Figures \ref{fig:teaser} and \ref{fig:comparison}, with the sharpened text embedding $\bfc_\phi$, \ourName achieves high-quality image generation comparable to guided sampling while requiring only half the computation. 

In text-to-image generation, descriptive text prompts are typically termed as positive prompts. However, images generated solely from positive prompts 
may not meet the desired quality standards. To address these issues, negative prompts are employed for image editing and quality enhancement. 
Previous works that distill CFG omit the entry for negative prompts, limiting practical applicability. 
In \ourName, we can integrate the embedding of negative text prompts $\bfc_n$ into the sharpener, which is achieved by $\bfc_\phi = \bfc + \alpha r_{\phi}(\bfc, \bfc_n) - \beta (\bfc_n-\bfc_{null})$, where $\beta$ is a hyperparamter controlling the strength of the introduced semantic shift. 
During training, negative prompts are randomly sampled from open-source datasets, and the training process remains consistent with Algorithm \ref{alg:alg}, except that negative text embeddings replace all the original null text embeddings. This strategy is especially effective for Stable Diffusion v1.5 variants~\cite{rombach2022ldm}, and we consider it an optional choice to endow \ourName sharpener with better robustness to the semantic shift (see Section \ref{sec:app_negative}).



\begin{figure}[t]
    \centering
    \includegraphics[width=\columnwidth]{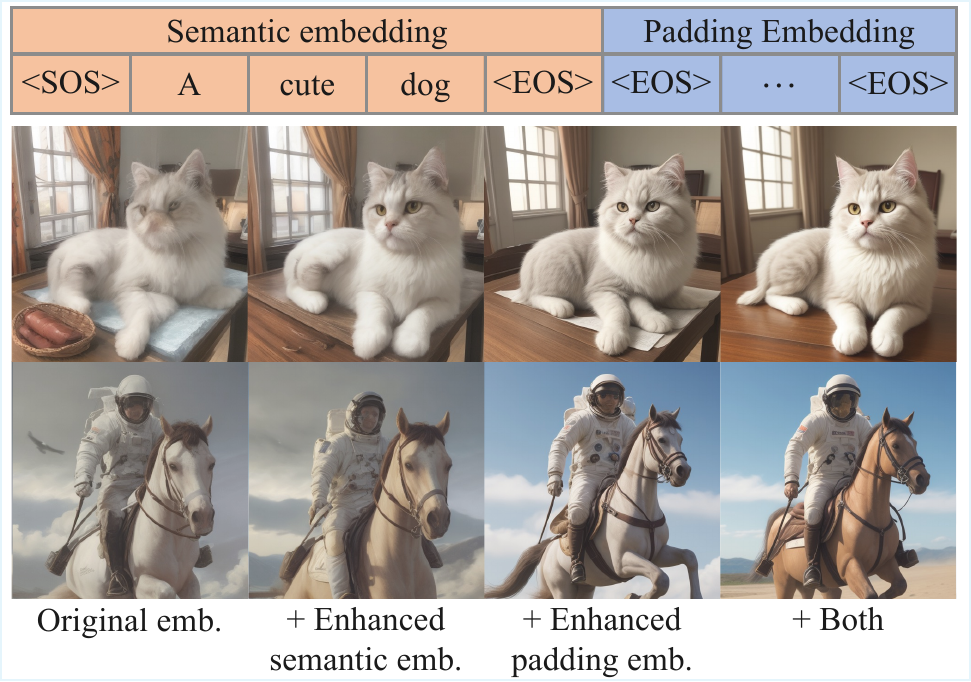}
    \caption{\textit{Top}: a text embedding consists of a semantic and padding embedding. \textit{Bottom}: replacing the original text embedding with the sharpened semantic and padding embedding. The latter largely improves the sample quality. 
    }
    \label{fig:emb_rep}
\end{figure}


\begin{table*}[ht]
    \centering
    \small
    \begin{tabular}{lrccccccccc}
        \toprule
        \multirow{2}{*}{Model} & \multirow{2}{*}{NFE} & \multirow{2}{*}{\# Param} & \multirow{2}{*}{FID ($\downarrow$)} & \multirow{2}{*}{CS ($\uparrow$)} & \multirow{2}{*}{AS ($\uparrow$)} & \multicolumn{4}{c}{HPS v2.1 ($\uparrow$)} & \multirow{2}{*}{DrawBench ($\uparrow$)} \\
        \cmidrule{7-10}
        & & & & & & Anime & Concept & Painting & Photo & \\
        \midrule
        \rowcolor[gray]{0.9} SD15 ($\omega=5$)           & 40 & - & 22.04 & 30.22 & 5.36 & 24.29 & 23.16 & 22.88 & 24.62 & 23.83 \\
        SD15 ($\omega=1$)           & 20 & - & 32.80 & 21.99 & 5.03 & 17.79 & 17.69 & 17.40 & 19.41 & 18.43 \\
        Scaling ($s=1.2$) $*$         & 20 & - & 32.54 & 22.89 & 5.13 & 18.11 & 17.94 & 17.73 & 19.57 & 18.80 \\
        GD$\dagger$~\cite{meng2023distillation}     & 20 & 859M & \underline{23.54} & 28.02 & \underline{5.30} & 21.84 & 20.58 & 20.19 & 23.48 & 21.99  \\
        PnP$\dagger$~\cite{hsiao2024plug}           & $\approx$ 28 & 361M & 26.57 & \underline{27.72} & \textbf{5.39} & \textbf{23.17} & \textbf{21.72} & \textbf{22.03} & \underline{24.17} & \underline{23.12}  \\
        \textbf{\ourName (ours)}    & 20 & 2M & \textbf{22.22} & \textbf{28.54} & 5.28 & \underline{22.78} & \underline{20.67} & \underline{20.71} & \textbf{24.96} & \textbf{23.32} \\
        \midrule
        \rowcolor[gray]{0.9} DreamShaper ($\omega=5$)    & 40 & - & 30.35 & 30.50 & 5.87 & 30.20 & 28.92 & 28.85 & 27.62 & 26.84 \\
        DreamShaper ($\omega=1$)    & 20 & - & \underline{24.17} & 27.22 & 5.74 & 24.42 & 24.44 & 24.61 & 23.56 & 22.05 \\
        Scaling ($s=1.3$) $*$         & 20 & - & \textbf{24.05} & 27.74 & 5.73 & 24.63 & 24.44 & 24.47 & 23.66 & 22.19 \\
        GD$\dagger$~\cite{meng2023distillation}     & 20 & 859M & 32.53 & \underline{28.48} & \underline{5.86} & 28.34 & 27.50 & 27.59 & 26.40 & 25.27 \\
        PnP$\dagger$~\cite{hsiao2024plug}           & $\approx$ 28 & 361M & 35.57 & 28.46 & \textbf{5.87} & \textbf{29.53} & \textbf{28.72} & \textbf{28.80} & \textbf{27.35} & \textbf{26.04} \\
        \textbf{\ourName (ours)}    & 20 & 2M & 30.36 & \textbf{29.03} & \textbf{5.87} & \underline{29.17} & \underline{28.44} & \underline{28.49} & \underline{27.27} & \underline{25.77} \\
        \midrule
        \rowcolor[gray]{0.9} SDXL ($\omega=5$)           & 40 & - & 23.95 & 32.10 & 5.60 & 29.67 & 28.19 & 28.19 & 26.51 & 26.03 \\
        SDXL ($\omega=1$)           & 20 & -  & 61.19 & 21.92 & 5.59 & 19.64 & 19.23 & 19.92 & 18.74 & 17.65 \\
        Scaling ($s=1.5$) $*$         & 20 & -  & 59.14 & 23.50 & 5.60 & 20.33 & 20.03 & 20.51 & 19.07 & 17.94 \\
        GD$\dagger$~\cite{meng2023distillation}     & 20 & 2.6B & \underline{28.88} & \textbf{30.84} & 5.57 & 28.83 & \underline{27.65} & \underline{28.11} & 26.39 & \underline{25.43} \\
        PnP$\dagger$~\cite{hsiao2024plug}           & $\approx$ 30 & 1.3B & 32.52 & 30.31 & \textbf{5.76} & \textbf{29.29} & 27.59 & \textbf{28.15} & \underline{26.44} & 25.35 \\
        \textbf{\ourName (ours)}    & 20 & 3M & \textbf{28.01} & \underline{30.63} & \underline{5.68} & \underline{29.06} & \textbf{27.72} & 28.10 & \textbf{26.48} & \textbf{25.44} \\
        \midrule
        \rowcolor[gray]{0.9} Pixart-$\alpha$ ($\omega=5$) & 40 & - & 38.39 & 30.67 & 6.03 & 31.43 & 29.97 & 29.60 & 28.97 & 27.95 \\
        Pixart-$\alpha$ ($\omega=1$) & 20 & - & 41.74 & 25.30 & \textbf{6.11} & 26.29 & 25.73 & 25.90 & 23.63 & 23.23 \\
        Scaling ($s=1.2$) $*$         & 20 & - & 41.89 & 25.79 & \underline{6.10} & 26.26 & 25.60 & 25.60 & 23.73 & 23.25 \\
        GD$\dagger$~\cite{meng2023distillation}     & 20 & 611M & 42.77 & 28.52 & 6.06 & 28.94 & 27.09 & 27.62 & \underline{26.68} & \underline{26.04} \\
        PnP$\dagger$~\cite{hsiao2024plug}           & $\approx$ 30 & 295M & \underline{40.06} & \textbf{29.55} & 5.99 & \underline{29.29} & \underline{28.24} & \underline{27.96} & 26.55 & 25.55 \\
        \textbf{\ourName (ours)}    & 20 & 5M & \textbf{39.80} & \underline{29.51} & 6.01 & \textbf{30.10} & \textbf{28.59} & \textbf{28.69} & \textbf{27.91} & \textbf{26.60} \\
        \bottomrule
    \end{tabular}
    \caption{Comparison of quantitative results.
    Images are generated with the same random seeds by the 20-step DPM-Solver++~\cite{lu2022dpmpp}. 
    $*$: Naive scaling using searched optimal scaling factor.
    $\dagger$: Our reimplementmentation of Guided Distillation (GD)~\cite{meng2023distillation} and Plug-and-Play Distillation (PnP)~\cite{hsiao2024plug}. PnP trains a ControlNet~\cite{zhang2023adding}, which introduces nearly half of the parameters of the base models and thus leads to larger NFE.
    }
    \label{tab:quantitative}
\end{table*}

\subsection{Inspecting the sharpened Text Embedding}
\label{sec:discussions}

Compared to existing works that distill CFG~\cite{meng2023distillation,hsiao2024plug}, decoupling the sharpener from the text-to-image model allows us to gain a deeper understanding of the proposed method by focusing on analyzing the sharpened text embeddings for inference. 
Next, we investigate the underlying mechanisms of our method and demonstrate how the sharpened text embeddings influence sample quality and sampling dynamics through both quantitative and qualitative evidence. 

The text embedding used for text-to-image generation consists of a $\textless SOS\textgreater$ token (start of sentence), some semantic tokens, and the remaining padded $\textless EOS\textgreater$ tokens (end of sentence).
As shown by previous works, e.g.,~\cite{yu2024uncovering}, based on the position of the first $\textless EOS\textgreater$ token, a text embedding can be divided into a semantic embedding that contains most semantic information and a padding embedding that encodes more about the image details. 
In Figure \ref{fig:emb_rep}, we replace the original embedding with sharpened semantic and padding embeddings. 
To replace the padding embedding, we recognize the index of the first $\textless EOS\textgreater$ token and then replace the embedding after this token with a sharpened one. It is observed that sharpened padding embeddings largely improve the image quality. Moreover, we compute the cosine similarity between $1,000$ paired original and sharpened semantic embeddings, obtaining a mean value of 0.75 and a standard deviation of 0.05, while for padding embeddings, they are 0.23 and 0.02. This indicates that padding embeddings are more significantly modified compared to semantic ones.
Combining both qualitative and quantitative results, we conclude that DICE mainly emphasizes sharpening the padding embedding while maintaining the original semantic embedding, leading to consistent semantic information but significantly improved image details.



%% file: secs/exp.tex
\section{Experiments}
\label{sec:exp}

\begin{figure*}[t]
\includegraphics[width=\textwidth]{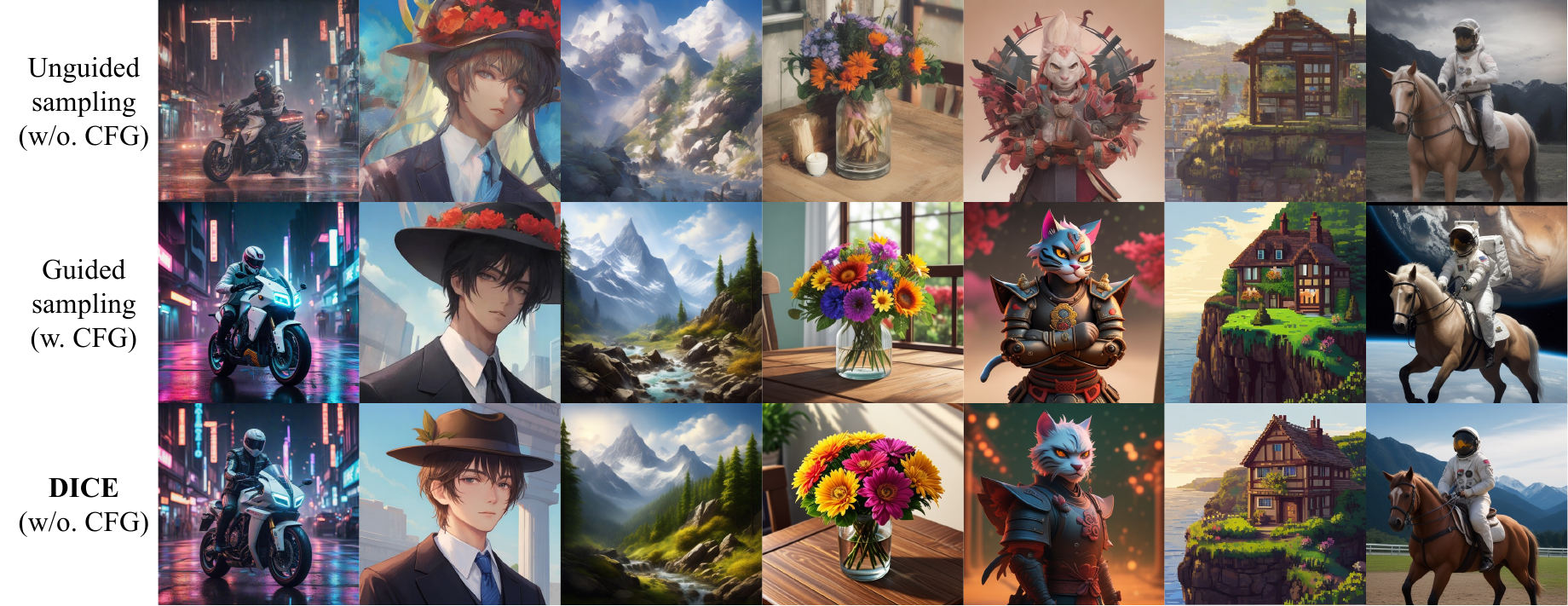}
\caption{Qualitative results with different model capacities, image styles, and network architectures. Images are generated by 20-step DPM-Solver++~\cite{lu2022dpmpp} on 7 text-to-image models including multiple SD15 variants~\cite{rombach2022ldm}, SDXL~\cite{podell2024sdxl} and Pixart-$\alpha$~\cite{chen2024pixart}. The prompts used are provided in Section \ref{sec:app_details}.}
\label{fig:qualitative}
\end{figure*}


\subsection{Text-to-Image Generation}
\label{sec:t2i}
\ourName's sharpener consists of two fully connected layers and an attention block. The number of trainable parameters is less than 1\% of the text-to-image model, leading to a negligible increase in computational overhead. The sharpening strength $\alpha=1$ and guidance scale $\omega=5$ are fixed during training. 
Our experiments are conducted on state-of-the-art text-to-image generation models, namely, Stable Diffusion v1.5 (SD15)~\cite{rombach2022ldm}, Stable Diffusion XL (SDXL)~\cite{podell2024sdxl}, Pixart-$\alpha$~\cite{chen2024pixart} and a series of SD15-based open source variants, 
including DreamShaper\footnote{\url{https://huggingface.co/Lykon/DreamShaper}}, 
AbsoluteReality\footnote{\url{https://huggingface.co/digiplay/AbsoluteReality_v1.8.1}},
Anime Pastel Dream\footnote{\url{https://huggingface.co/Lykon/AnimePastelDream}},
DreamShaper PixelArt\footnote{\url{https://civitai.com/models/129879/dreamshaper-pixelart}}, 
and 3D Animation Diffusion\footnote{\url{https://civitai.com/models/118086?modelVersionId=128046}}. 
We use MS-COCO 2017~\cite{lin2014microsoft} for training and evaluation.
More details about training, evaluation, and pre-trained models are included in Section \ref{sec:app_details}.

We evaluate \ourName on text-to-image models with varying capacities, ranging from 0.6B to 2.6B parameters, across architectures such as U-Net~\cite{ronneberger2015u} and DiT~\cite{peebles2023scalable}, and across diverse image styles including dreamlike, realistic, 3-D, pixel art, and anime style. 
The sample quality are measured by the Fr\'echet Inception Distance (FID)~\cite{heusel2017gans}, CLIP Score (CS)~\cite{radford2021learning}, Aesthetic Score (AS)~\cite{schuhmann2022laion}, HPS v2.1~\cite{wu2023human} and DrawBench~\cite{saharia2022photorealistic}.
Quantitative results are presented in Table \ref{tab:quantitative}. Our enhanced unguided sampling achieves sample quality comparable to that of guided sampling and largely outperforms the original unguided sampling, as illustrated in Figure \ref{fig:qualitative}. Moreover, with only the text embedding modified, \ourName achieves performance comparable to existing methods~\cite{meng2023distillation,hsiao2024plug} with largely reduced trainable hyperparameters and without increasing inference costs.

\subsection{Discussion and Ablation Study}
\label{sec:discussion}

\begin{figure*}[t]
    \includegraphics[width=\textwidth]{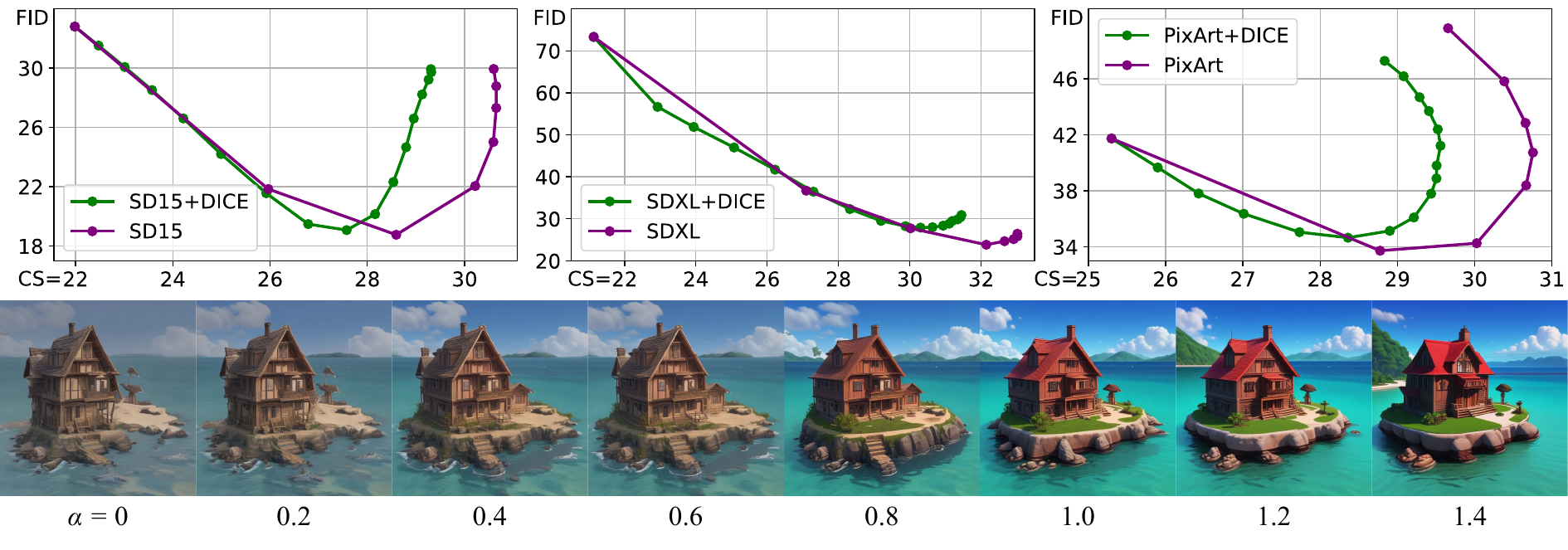}
    \captionsetup{skip=2pt}
    \caption{\small \textit{Top}: FID-CS curves over guidance scale (for guided sampling) and sharpening strength $\alpha$ (for \ourName) on different text-to-image models. The sharpening strength acts like the guidance scale. Guidance scales: $\{1, 1.5, 2.5, 5, 7.5, 10, 12.5, 15\}$. Sharpening strengths: $\{0, 0.1, 0.2, \cdots, 1.6\}$. \textit{Bottom}: as $\alpha$ increases, the sharpener can maintain the semantic information while improving the sample quality. 
    }
    \label{fig:fid_cs}
\end{figure*}

\textbf{Sharpening strength $\alpha$.}
In practical applications, CFG offers flexibility in controlling image quality by adjusting the guidance scale. Although \ourName maintains a fixed guidance scale during training, it allows for this flexibility via the sharpening strength $\alpha$. This capability stems from the underlying mechanism of \ourName, which emphasizes enhancing image details while preserving semantic information. Figure \ref{fig:fid_cs} presents a comprehensive evaluation, demonstrating that the sharpening strength $\alpha$ serves a role akin to that of the guidance scale $\omega$. 


\textbf{Generalization.} As the sharpener operates independently of the text-to-image model, we investigate the feasibility of applying a well-trained sharpener to unseen text-to-image models and text prompts. In Figure \ref{fig:generalization}, we separately train three sharpeners (sharpener $i$, $i=1,2,3$) on three distinct text-to-image models, i.e., DreamShaper (model $1$), DreamShaper PixelArt (model $2$), and Anime Pastel Dream (model $3$). Subsequently, we plug each sharpener into all models for unguided image generation. The results show that the sharpeners exhibit strong generalization capabilities across diverse domains, consistently and significantly improving the original unguided sampling. 
In Figure \ref{fig:generalization_prompts}, we further investigate the generalization ability of \ourName on unseen prompts outside the training dataset. We test the performance of \ourName on unusual and long prompts and find \ourName closely mimics the behavior of guided sampling and generalizes well to challenging text prompts.


%% file: secs/related.tex
\section{Related Works}
\label{related}



\textbf{CFG distillation.} 
Previous works have proposed distilling CFG-based text-to-image models. Guided distillation~\cite{meng2023distillation} incorporates the guidance scale as a new model input through fine-tuning, a method later adopted by FLUX\footnote{\url{https://github.com/black-forest-labs/flux}}. Plug-and-Play~\cite{hsiao2024plug} trains an auxiliary guided model attached to the U-Net decoder, which is transferable to new domains. A recent work, NoiseRefine~\cite{ahn2024noise}, proposes to refine the initially sampled Gaussian noise to enhance unguided sampling. However, the distillation loss requires samples generated by both unguided and guided sampling, introducing extensive computational overhead during training. In contrast, our method solely modifies the text conditioning without altering the generative process of diffusion models and retains fast training speed. A more detailed comparison is provided in Section \ref{sec:comparison_w_distillation}.

\textbf{Reward-based methods.} The text encoder plays a crucial role in text-to-image generation. Several studies aim to improve guided sampling by fine-tuning the text encoder through reinforcement learning~\cite{chen2024enhancing} and reward propagation~\cite{li2024textcraftor}. Our method differs in two key aspects. First, it is specifically designed to improve unguided sampling without relying on CFG. Second, it is trained under CFG-based supervision and does not require human feedback or any reward models. We include further discussion in Section \ref{sec:comparison_w_reward}.

\begin{figure}[t]
    \centering
    \includegraphics[width=0.98\columnwidth]{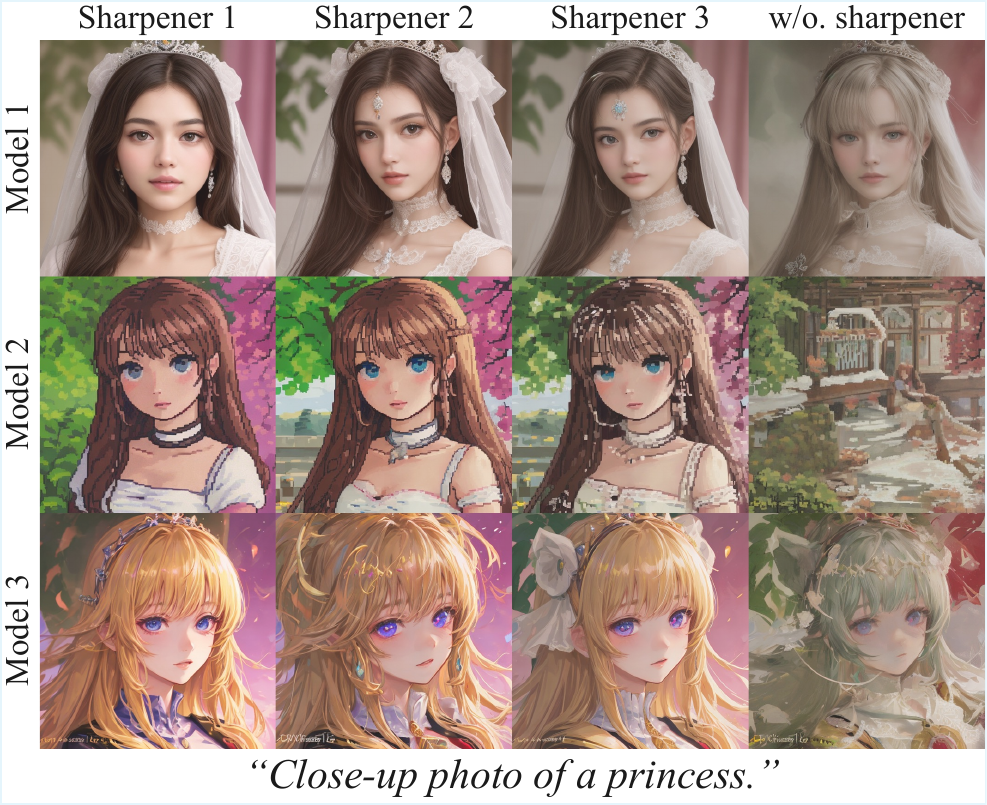}
    \caption{Generalization across different models. The original unguided sampling results are provided for comparison. Each sharpener is independently trained on DreamShaper (model $1$), DreamShaper PixelArt (model $2$), or Anime Pastel Dream (model $3$), and applied to other models. 
    }
    \label{fig:generalization}
\end{figure}

\begin{figure}[t]
    \centering
    \includegraphics[width=0.98\columnwidth]{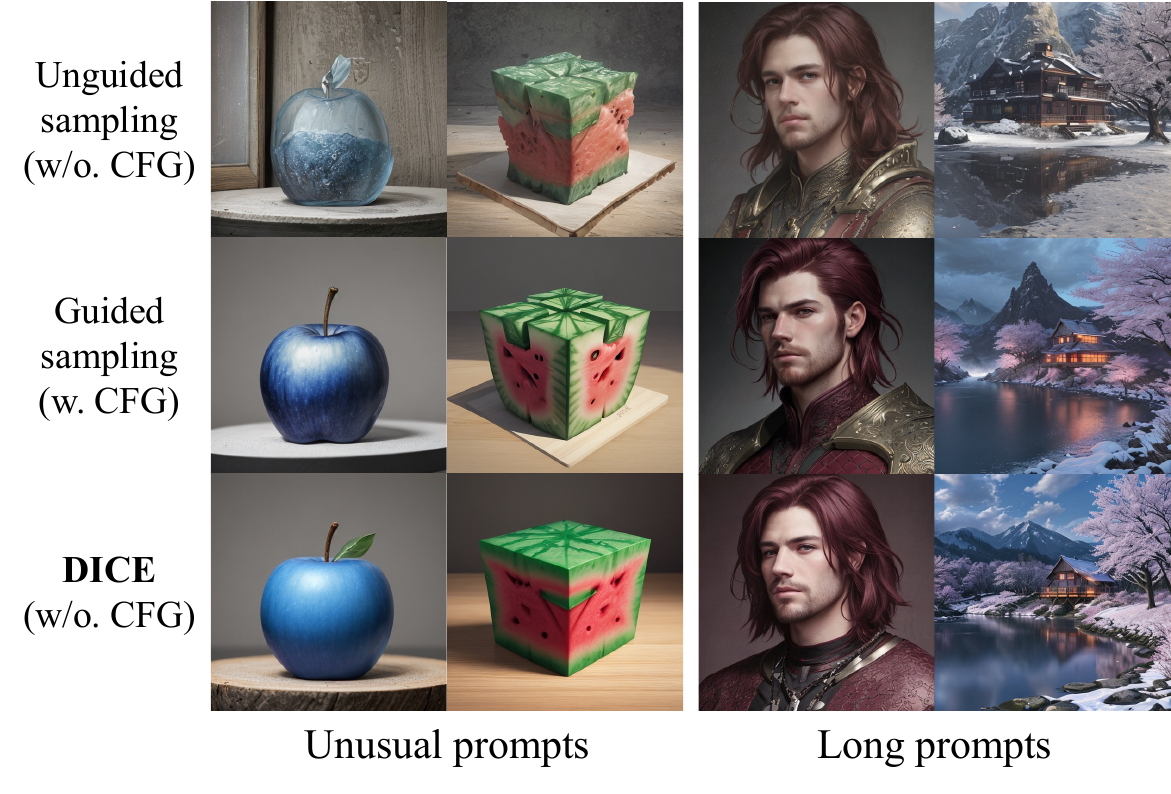}
    \caption{Generalization of \ourName to unusual and long text prompts. \ourName closely mimics the behavior of guided sampling and generalizes well to unseen text prompts. The unusual prompts are ``\textit{A blue apple}'' and ``\textit{A cubic watermelon}''. The detailed long prompts are provided in Section \ref{sec:app_details}.}
    \label{fig:generalization_prompts}
\end{figure}

%% file: secs/conclusion.tex
\section{Conclusion}
\label{sec:conclusion}
Classifier-Free Guidance (CFG) is a prevalent technique in text-to-image generation, enhancing image quality but introducing increased sampling overhead. In this work, we introduce \ourName, which fortifies text embeddings by training a sharpener under CFG-based supervision, achieving efficient and effective unguided text-to-image generation. 
We reveal that \ourName enhances fine-grained image details through a universal enhancement pattern without compromising essential semantic information. 
Extensive experiments across various model capacities, image styles, and architectures demonstrate the effectiveness of our method. 
Our approach also exhibits strong generalization capability on unseen text-to-image models and challenging text prompts. 

\textbf{Limitations.} Similar to existing methods that distill the CFG, the performance of \ourName has not yet converged to the level of guided sampling. To overcome this limitation, future work will focus on enhancing our method beyond guided sampling by mitigating the information loss caused by distillation. 
Exploring ways to improve unguided sampling without CFG-based supervision is also a promising direction. 


%% file: secs/appendix.tex
\appendix
\clearpage
\setcounter{secnumdepth}{2}
\renewcommand{\thesubsection}{\Alph{section}.\arabic{subsection}}

\section{Appendix}
\label{sec:app}

\subsection{Additional Details}
\label{sec:app_details}

\textbf{Sharpener design.} For the network design of \ourName sharpeners, we stack two fully-connected layers and one cross-attention layer. The first fully-connected layer compresses the input two text embeddings into a context dimension of 512. The obtained two features are then input to the cross-attention layer and are finally extended to the original context dimension through the second fully-connected layer. The inner context dimension of 512 is set to control the total number of parameters of the sharpener.
The number of parameters of the obtained sharpener accounts for only 0.21\%, 0.12\% and 0.86\% of SD15 variants, SDXL and PixArt-$\alpha$, respectively.
The extra sampling overhead of the sharpener is negligible since it only operates once for every generation. 

\textbf{Training.} During the training of \ourName, we use L1 norm as loss function. For optimization, we use the Adam optimizer~\cite{kingma2014adam} with $\beta_1=0.9$, $\beta_2=0.999$, a learning rate of $2e-4$, and a batch size of $128$. 
Sharpeners for SD15-based variants are trained with negative prompts with $\beta=0.5$. 
All models are trained with $\sim 8,000$ gradient updates, requiring 4.5, 6, and 18 hours for the SD15-based variants, Pixart-$\alpha$ and Stable Diffusion XL, respectively, using 8 NVIDIA A100 GPUs.

\textbf{Evaluation.} The sample quality is evaluated by the Fr\'echet Inception Distance (FID)~\cite{heusel2017gans}, CLIP Score (CS)~\cite{radford2021learning}, Aesthetic Score (AS)~\cite{schuhmann2022laion}, HPS v2.1~\cite{wu2023human} and DrawBench~\cite{saharia2022photorealistic}. To compute FID, we generate $5,000$ images using $5,000$ prompts sampled from the MS-COCO 2017 validation set and use the validation set as reference images. The $5,000$ generated images are also used to compute CS and AS.

\textbf{Prompts.}
We present the text prompts and models used to generate Figure \ref{fig:qualitative} and Figure \ref{fig:generalization_prompts} in Table \ref{tab:quantitative_setting} and Table \ref{tab:generalization_prompts}. They are selected to generate images with as many image styles and topics as possible.

\textbf{Pre-trained models.}
In Table \ref{tab:model_setting}, we summarize all the text-to-image models used in our experiments. SD15-based variants~\cite{rombach2022ldm} and SDXL~\cite{podell2024sdxl} use U-Net~\cite{ronneberger2015u} as backbone while PixArt-$\alpha$~\cite{chen2024pixart} uses DiT~\cite{peebles2023scalable}. Different pre-trained text encoders are used for each type of text-to-image model with the number of parameters ranging from different orders of magnitude.

\begin{table*}[!htbp]
\centering
\begin{tabular}{p{4cm}p{8cm}}
\toprule
Model & Text prompt \\
\midrule
SDXL~\cite{podell2024sdxl}  & \textit{A rainy street, a racer on a white motorcycle by the street, bright neon lights, cyberpunk style, futuristic, 8k, best quality, clear background} \\
Anime Pastel Dream & \textit{A man in suits and hat, center, close-up, best quality} \\
Pixart-$\alpha$~\cite{chen2024pixart} & \textit{Epic scene, mountains, sunshine, trees, rocks, clear, realistic, best quality, best detail, aesthetic, masterpiece} \\
AbsoluteReality & \textit{Colorful flowers in a vase on a wooden table, sunshine, aesthetic, realistic, 8k, best quality} \\
3D Animation Diffusion & \textit{An anthropomorphic cat samurai wearing armor, bokeh temple background, colorful, masterpieces, best quality, aesthetic} \\
DreamShaper PixelArt & \textit{A countryside cottage on the edge of a cliff overlooking an ocean, pixel art} \\
DreamShaper & \textit{Photo of an astronaut riding a horse} \\
\bottomrule
\end{tabular}
\caption{Text-to-image models and text prompts used in Figure \ref{fig:qualitative}.}
\label{tab:quantitative_setting}
\end{table*}

\begin{table*}[!htbp]
\centering
\begin{tabular}{p{4cm}p{8cm}}
\toprule
Model & Text prompt \\
\midrule
DreamShaper & \textit{Realistic portrait of a man, masculine face, medium hair, Maroon hair, masculine, athletic, intricate details on clothing, perfect composition, deviant art hd, art station hd, concept art, detailed face and body, award-winning photography, detailed face} \\
DreamShaper & \textit{Anti-burn, no mist, photorealistic, 8k, best render, render, winter, nighttime, cloud, cherry blossom, day, fantasy, fish, lake, landscape, high snowy mountain, no humans, ocean, outdoors, river, scenery, sky, splashing, water, watercraft, waterfall, waves, ultra realistic, photorealistic, sea} \\
\bottomrule
\end{tabular}
\caption{Text-to-image models and text prompts used in Figure \ref{fig:generalization_prompts}.}
\label{tab:generalization_prompts}
\end{table*}

\begin{table*}[!htbp]
\begin{center}
\begin{tabular}{lccc}
\toprule
Model & SD15 variants & SDXL & PixArt-$\alpha$ \\
\midrule
Model architecture          & U-Net & U-Net & DiT   \\
\# of model parameters      & 0.86B & 2.58B & 0.61B \\
Text encoder                & CLIP ViT-L & CLIP ViT-L \& OpenCLIP ViT-bigG & Flan-T5-XXL \\
\# of tokens                & 77    & 77    & 120   \\
Context dimension           & 768   & 2048  & 4096  \\
\# of encoder parameters    & 0.12B & 0.82B & 4.76B \\
\# of sharpener parameters   & 1.84M & 3.15M & 5.25M \\
\bottomrule
\end{tabular}
\end{center}
\caption{Summary of used text-to-image models.}
\label{tab:model_setting}
\end{table*}

\subsection{Negative Prompts}
\label{sec:app_negative}
\begin{figure*}[t]
    \begin{subfigure}[t]{0.475\textwidth}
        \includegraphics[width=\textwidth]{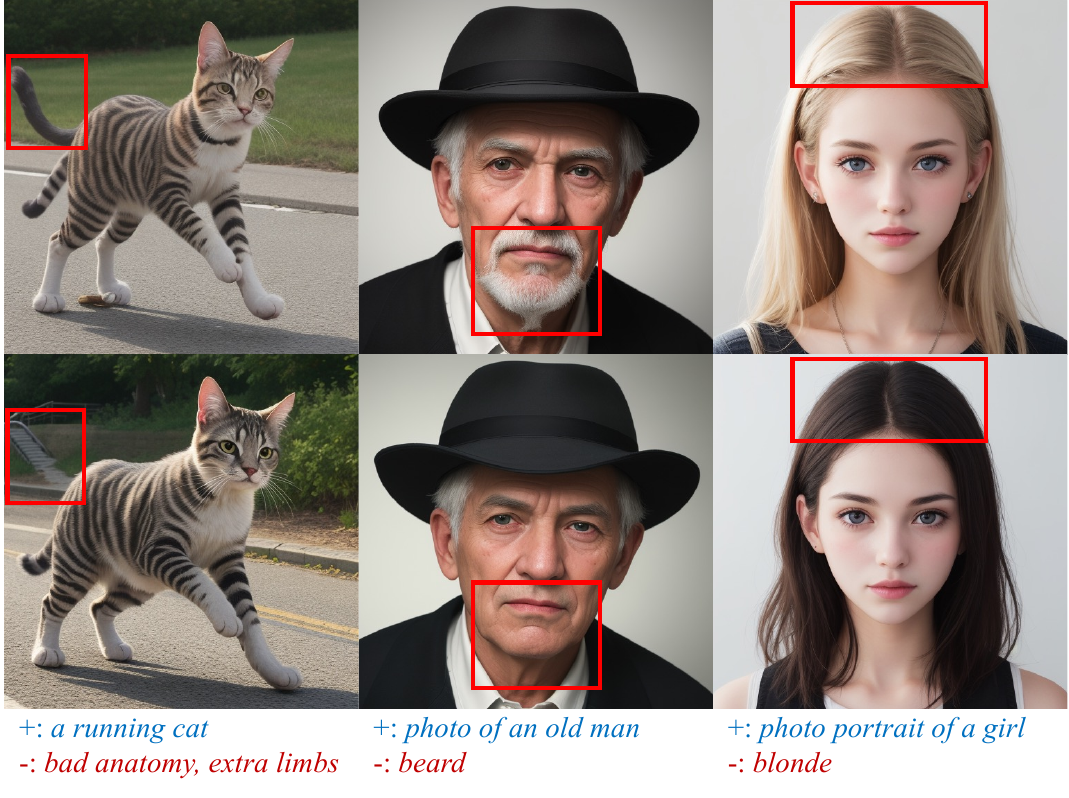}
    \end{subfigure}
    \hfill
    \begin{subfigure}[t]{0.495\textwidth}
        \includegraphics[width=\textwidth]{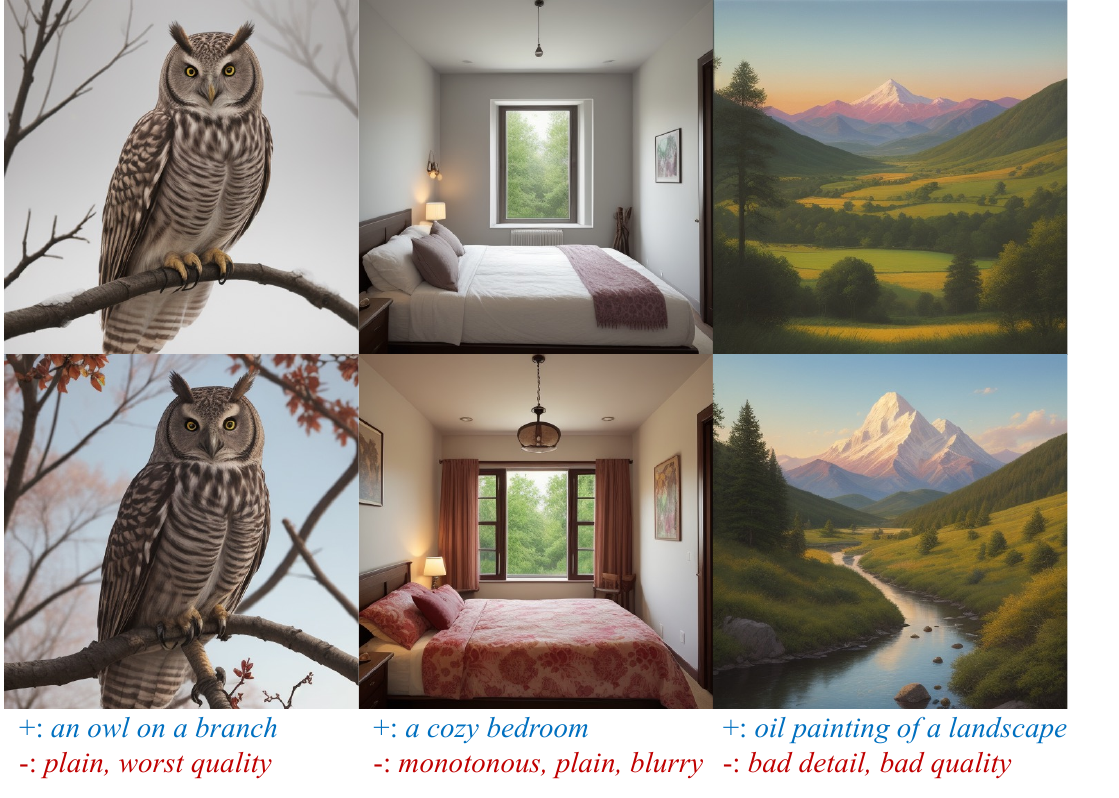}
    \end{subfigure}
    \caption{Performance of \ourName on negative prompts for image editing (left) and quality improvement (right). Positive and negative prompts are denoted by $+$ and $-$, respectively.}
    \label{fig:negative}
\end{figure*}

As mentioned in Section \ref{sec:dice}, \ourName is endowed with the ability to integrate negative prompts.
In Figure \ref{fig:negative}, we show the effectiveness of \ourName on the two main purposes of using negative prompts.
Our method can perform desirable image editing and quality improvement, including modifying unnatural limbs, removing or changing unwanted features, and handling abstract prompts related to image quality.

In Table \ref{tab:negative}, we provide additional results for negative prompts during inference using SD15 with both baseline (guided sampling with $\omega = 5$) and \ourName. Negative prompts are randomly sampled for each image. While using negative prompts improves visual quality, it may cause a worse FID score.

\subsection{Comparison with Distillation-based Methods}
\label{sec:comparison_w_distillation}
\begin{figure*}[t]
\begin{center}
    \begin{subfigure}[t]{0.325\textwidth}
        \includegraphics[width=\textwidth]{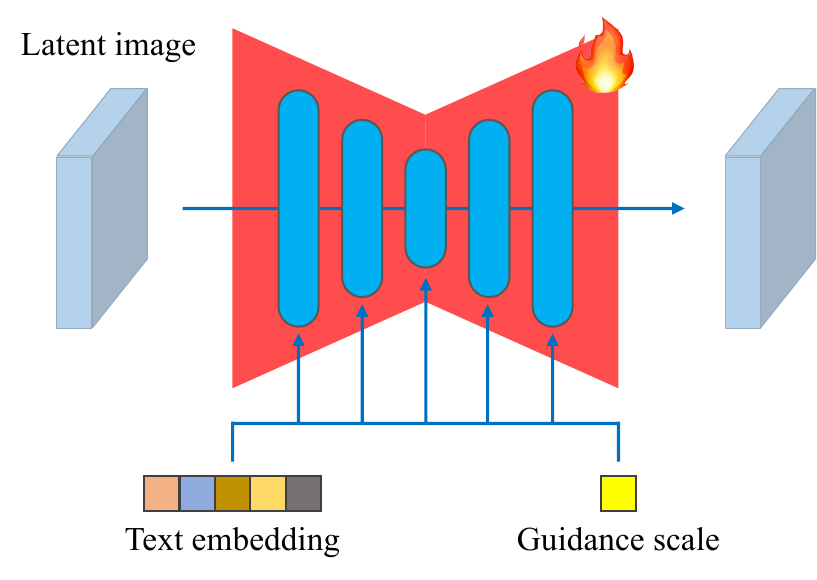}
        \caption{Guidance Distillation~\cite{meng2023distillation}.}
    \end{subfigure}
    \hfill
    \begin{subfigure}[t]{0.325\textwidth}
        \includegraphics[width=\textwidth]{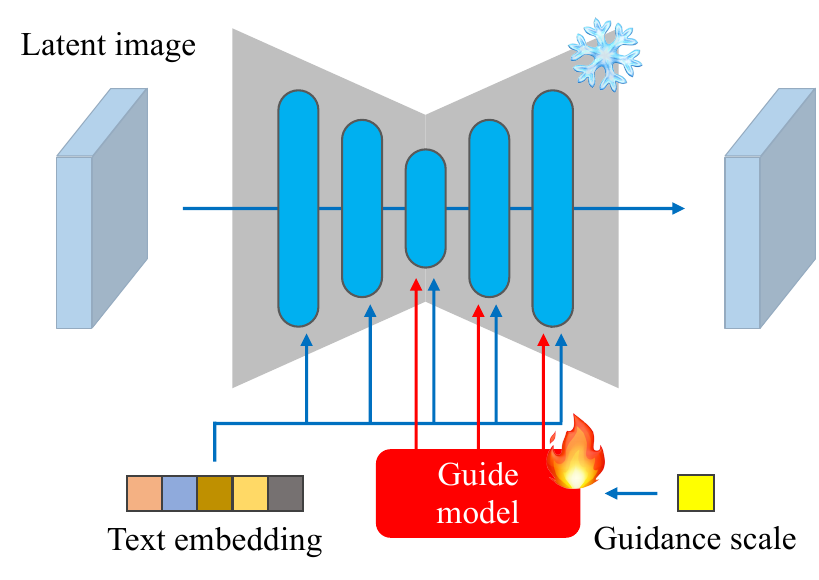}
        \caption{Plug-and-Play~\cite{hsiao2024plug}.}
    \end{subfigure}
    \hfill
    \begin{subfigure}[t]{0.325\textwidth}
        \includegraphics[width=\textwidth]{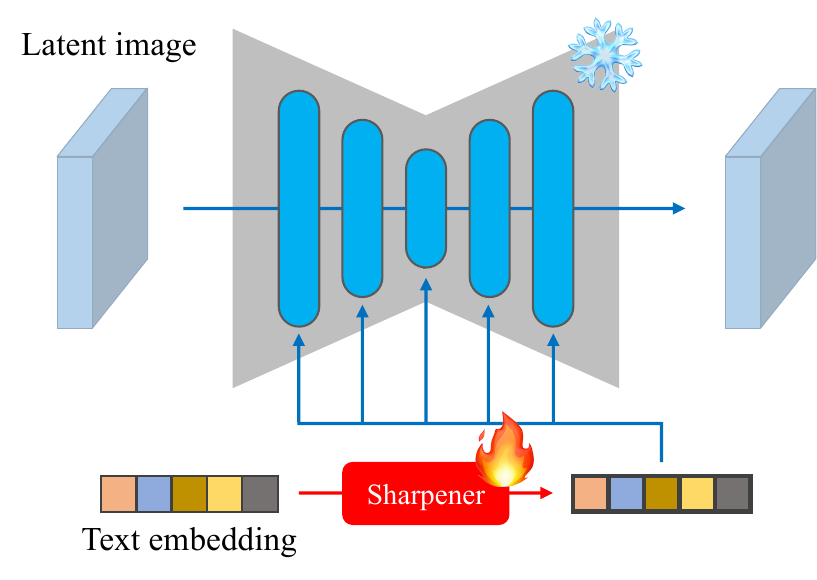}
        \caption{Our method.}
    \end{subfigure}
    \caption{Method comparison. (a) Guidance Distillation takes the guidance scale as an additional entry and fine-tunes the whole text-to-image model. The way to process the guidance scale is similar to that of the timestamp. (b) Plug-and-Play takes inspiration from ControlNet~\cite{zhang2023adding} and trains an external guide model to distill the guidance scale. (c) Our method trains an sharpener to enhance the text embedding. We completely decouple the sharpener from the text-to-image model, which is extremely easy to implement and exhibits better interpretability.}
    \label{fig:comparison_w_distillation}
\end{center}
\end{figure*}

The most related works to ours are Guidance Distillation~\cite{meng2023distillation} and Plug-and-Play~\cite{hsiao2024plug}, which also aim at reducing the sampling overhead of CFG through distillation. In Figure \ref{fig:comparison_w_distillation}, we provide an illustrative comparison between them and our method. 
Guidance Distillation takes the guidance scale as an additional model entry and processes it in the way similar to the timestamp. The whole parameters of the text-to-image model are fine-tuned under CFG-based supervision. Plug-and-Play trains a guide model in a similar way, where the guide model interacts with the intermediate features of the text-to-image model as in ControlNet~\cite{zhang2023adding}. Our method, instead, completely decouples the sharpener from the text-to-image model by only modifying the text embedding, which is essentially the model condition. Despite a small number of trainable parameters, our method achieves comparable performance with Guidance Distillation as verified in Section \ref{sec:t2i}. This decoupling further enhances the interpretability of our method. Moreover, \ourName is easier to deploy because both Guidance Distillation and Plug-and-Play are associated with specific layers in the text-to-image model, while DICE only requires the output of the text encoder, which can be easily accessed externally from the text-to-image model. For example, integrating \ourName sharpener into the highly encapsulated toolkit \textit{diffusers} requires only adding five lines of code to the inference pipeline.

\begin{table}[t]
    \centering
    \begin{tabular}{lccc}
        \toprule
        Method & FID & CS & AS \\
        \midrule
        Baseline w/o. negative prompts       & 22.06 & 30.23 & 5.36\\
        Baseline w/.  negative prompts        & 24.48 & 28.80 & 5.43 \\
        \ourName w/o.  negative prompts       & 22.22 & 28.54 & 5.28 \\
        \ourName w/.  negative prompts        & 22.42 & 28.64 & 5.29 \\
        \bottomrule
    \end{tabular}
    \caption{The effect of negative prompts during inference.}
    \label{tab:negative}
\end{table}

\subsection{Comparison with Reward-based Methods}
\label{sec:comparison_w_reward}
\begin{figure*}[t]
\begin{center}
    \begin{subfigure}[t]{\textwidth}
        \includegraphics[width=\textwidth]{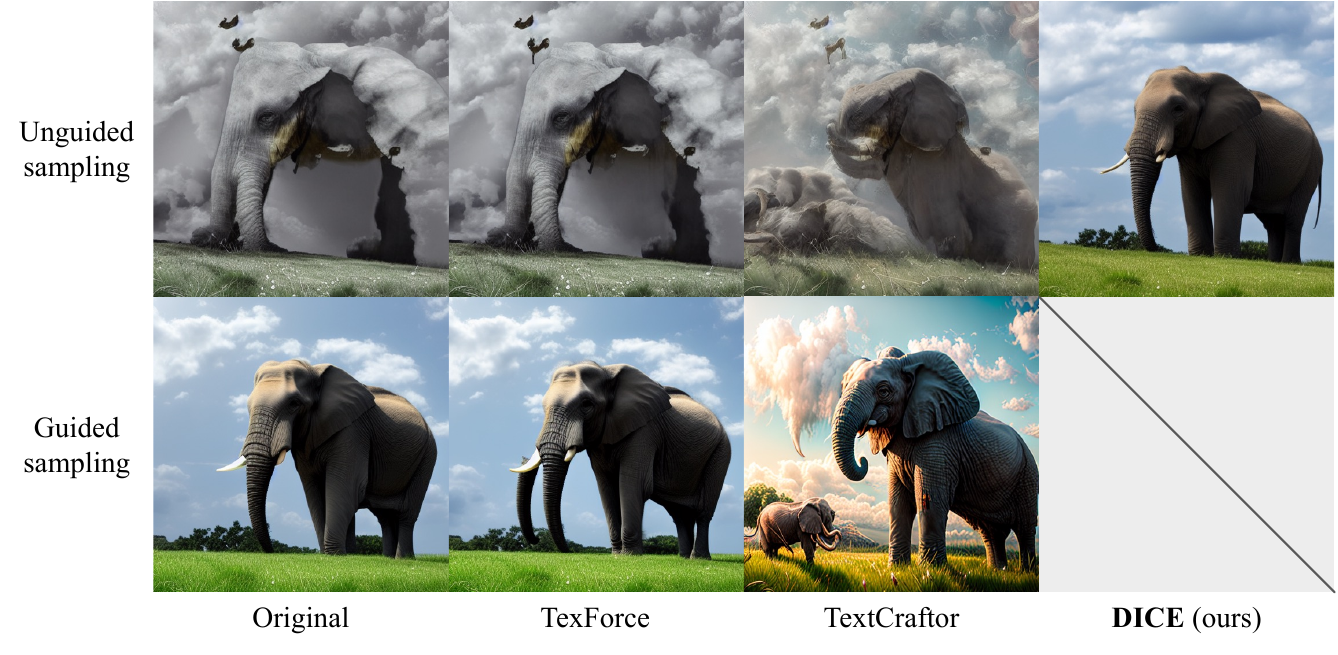}
        \caption{Model: Stable Diffusion v1.5~\cite{rombach2022ldm}. Text prompt: ``\textit{an elephant on the grassland, cloud, high quality, realistic}''.}
    \end{subfigure}
    \vskip 0.15in
    \begin{subfigure}[t]{\textwidth}
        \includegraphics[width=\textwidth]{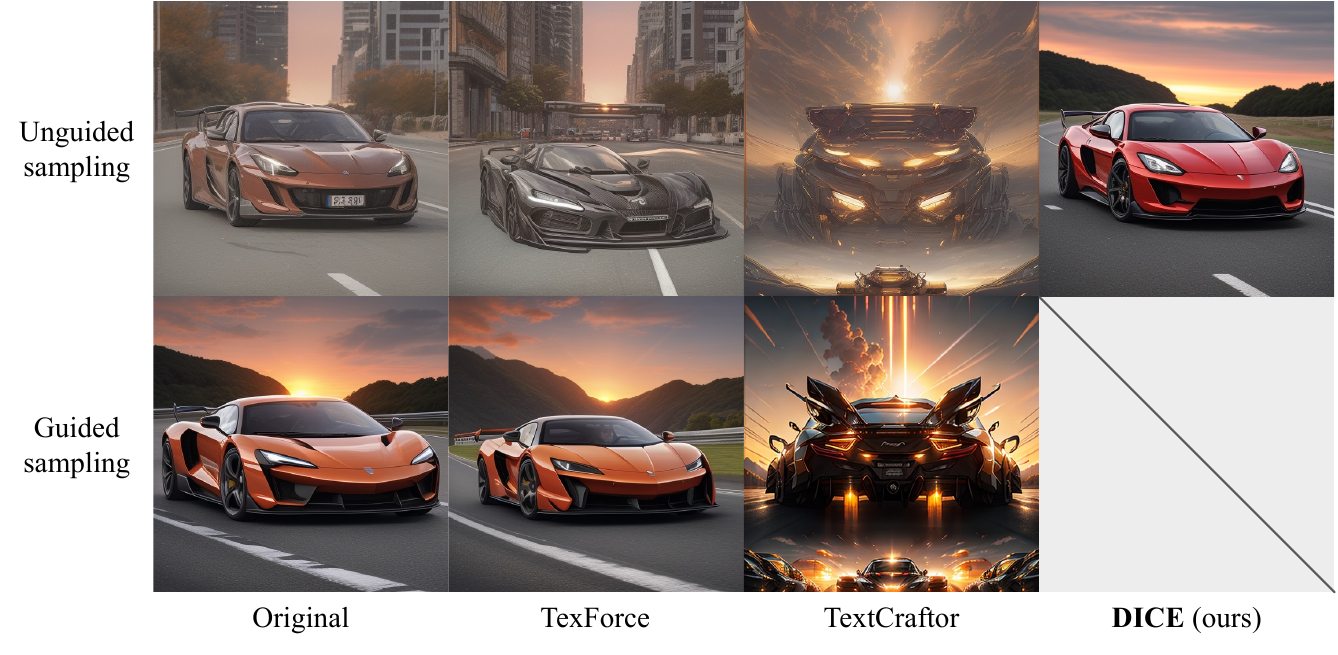}
        \caption{Model: DreamShaper. Text prompt: ``\textit{a supercar on the road, sunset, high quality}''.}
    \end{subfigure}
\caption{Comparison with reward-based methods, i.e., TexForce~\cite{chen2024enhancing} and TextCraftor~\cite{li2024textcraftor}. The text embeddings obtained by reward-based methods are not applicable to unguided sampling. \ourName achieves high-quality unguided sampling with reduced sampling overhead.}
\label{fig:app_comp}
\end{center}
\end{figure*}

Aiming at further enhancing the image quality given by guided sampling, previous works have proposed to fine-tune the text encoder through reinforcement learning~\cite{chen2024enhancing} and reward propagation~\cite{li2024textcraftor}. In Figure \ref{fig:app_comp}, we provide a qualitative comparison between our method and these reward-based methods, i.e., TexForce~\cite{chen2024enhancing} and TextCraftor~\cite{li2024textcraftor}. Though these reward-based methods improve the sample quality of guided sampling, the text embeddings are not applicable to unguided sampling. We re-train TexForce for unguided sampling but only observe minor improvement. Therefore, the mechanism of our method, as illustrated in Section \ref{sec:discussions}, is different from that of reward-based methods, which we hypothesize is due to the direct CFG-based supervision instead of reward models.

\subsection{Distributional Characteristics of Sharpened Text Embeddings}
\label{sec:app_visulization}

\begin{figure*}[t]
    \centering
    \includegraphics[width=\textwidth]{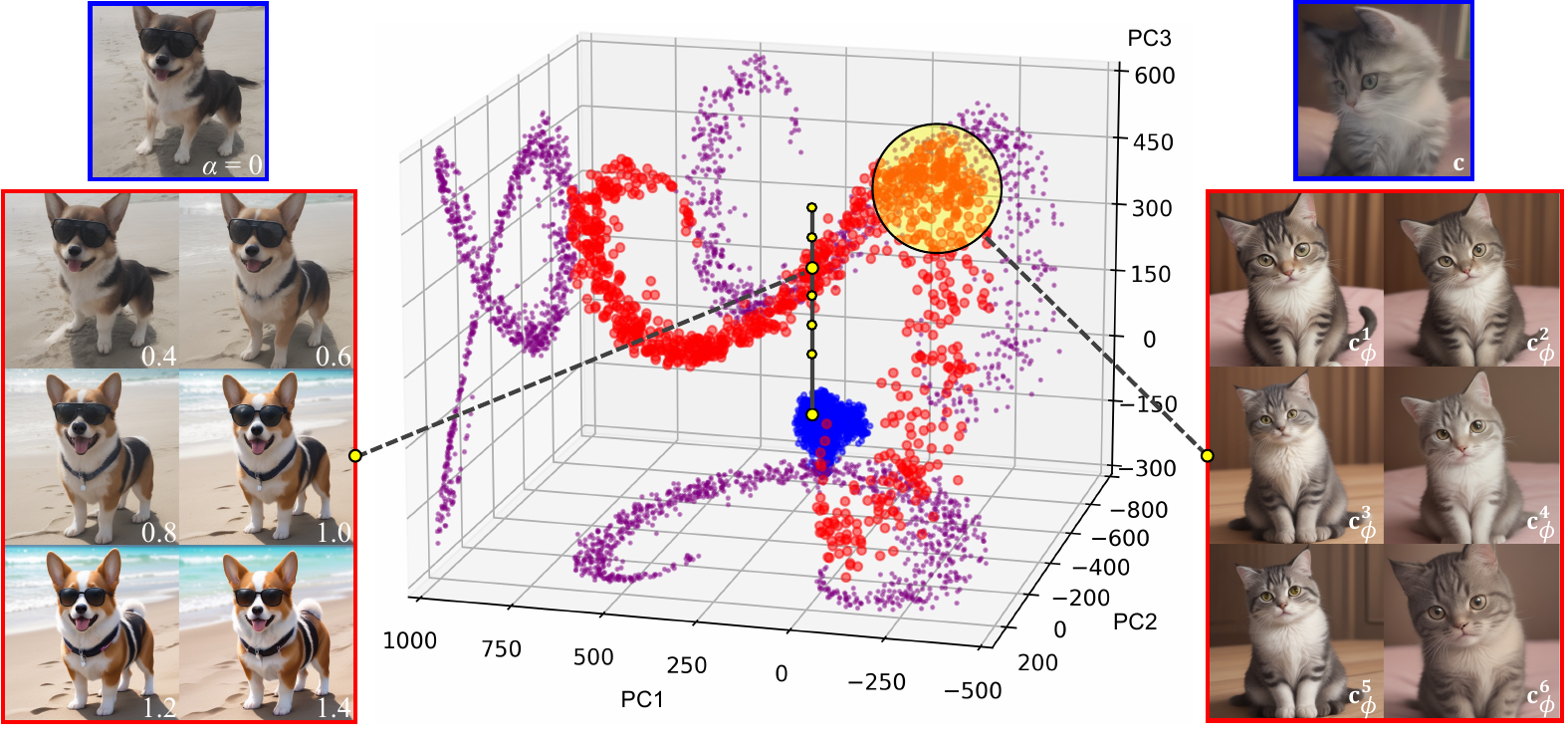}
    \caption{Visualization of 1,000 text embeddings through PCA. Blue scatters are the original text embeddings. The sharpened text embeddings (red scatters) are distributed on a simplified manifold. Left: images generated by ablating the sharpener strength $\alpha$ with prompt ``\textit{A corgi wearing sunglasses on the beach}''. Right: images generated by $\bfc_\phi^i=\bfc+r_\phi(\bfc_i^*,\bfc_\emptyset)$ where $i=1,\cdots,6$ and $\bfc$ is given by ``\textit{a cute cat, perfect detail, best quality}''. $\bfc_i^*$s are respectively encoded by the original prompt, ``\textit{photo portrait of a girl}'', ``\textit{a cozy bedroom}'', ``\textit{xjhgbion}'', ``\textit{2!0@2$\#$5}'', and a null text, indicating different red scatters on the manifold. 
    }
    \label{fig:pca_main}
\end{figure*}

As \ourName focuses on enhancing image details—which influence the overall preferences and are less relevant to text—we aim to investigate the distributional characteristics of the resulting sharpened text embeddings. In Figure \ref{fig:pca_main}, we visualize 1,000 text embeddings with a dimension of $59,136$ (77 $\times$ 768) through standard principal component analysis (PCA). The sharpened text embeddings (red scatters) are regularly distributed on a simplified manifold.

On the left side of Figure \ref{fig:pca_main}, we examine the effect of the sharpening strength $\alpha$, which is fixed to 1 during training, and $\alpha=0$ corresponds to the original unguided sampling. It is shown that the role of $\alpha$ is similar to that of the guidance scale. With the increase of $\alpha$, the generated image improves with richer details and stronger contrast while preserving consistent semantic information. 
On the right side, we move the original text embedding (e.g., corresponding to a cat) to different positions on the red manifold by providing the sharpener with different inputs. It is shown that the image details are always improved, even when the input to the sharpener is irrelevant to the original text (e.g., a random combination of letters ``\textit{xjhgbion}'' is completely irrelevant to a cat). This indicates that \ourName identifies a universal enhancement pattern that maintains the original semantic information while strengthening image details. 
We note that these observations are made possible due to the complete decoupling of our sharpener and the text-to-image model.

\begin{figure*}[t]
\begin{center}
    \includegraphics[width=\textwidth]{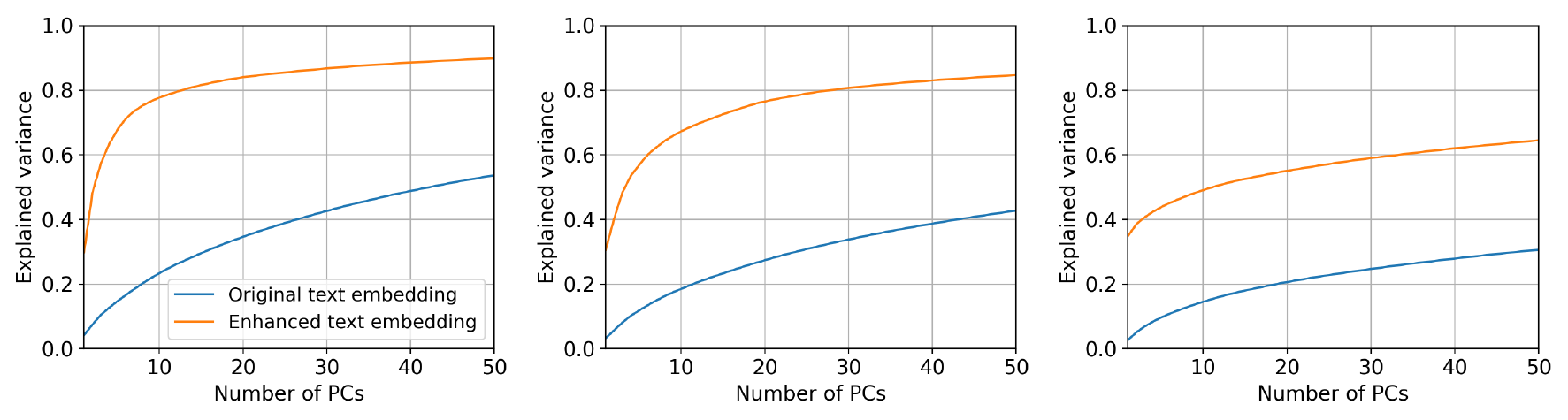}
    \caption{Explained variance with respect to the number of principal components (PCs) on SD15 (left), SDXL (middle) and PixArt-$\alpha$ (right) for both the original and sharpened text embeddings. 
    We calculate the explained variance as the ratio of the sum of the top squared eigenvalues to the sum of all squared eigenvalues. 
    The explained variance given by sharpened text embeddings is considerably larger than that of original embeddings, due to the existence of the universal enhancement pattern.
    }
    \label{fig:pca_ratio}
\end{center}
\end{figure*}

To quantify the degree to which our \ourName sharpener simplifies the original complex manifold, we calculate the explained variance by the top principal components (PCs), which is given by the ratio of the sum of the top squared eigenvalues to the sum of all squared eigenvalues. The results are shown in Figure \ref{fig:pca_ratio}. The universal enhancement pattern implied in the sharpened text embeddings largely simplifies the original manifold, which is indicated by the considerably larger explained variance compared to that given by the original text embeddings.

\subsection{Additional Discussions and Ablation Studies}
\label{sec:app_ablation}

\textbf{Sharpener design.} \ourName sharpener consists of two fully connected layers and one cross-attention layer. In Table \ref{tab:architecture}, we show the performance of \ourName with alternative network designs and conclude that the performance of \ourName is insensitive to the choice of sharpener architecture. 

\begin{table}[t]
    \centering
    \begin{tabular}{lcccc}
        \toprule
        Architecture & $\#$ params & FID & CS & AS \\
        \midrule
        \ourName & 1.8M & 27.57 & 29.20 & 5.72 \\
        MLP & 1.1M & 29.25 & 29.25 & 5.75 \\
        Transformer & 11.6M & 30.89 & 29.14 & 5.79 \\
        \bottomrule
    \end{tabular}
    \caption{Ablation study on sharpener design with DreamShaper. We provide additional results on sharpeners built by a 3-layer MLP and a 2-layer transformer encoder. Sharpeners are all trained with ~800 gradient updates. The performance of \ourName is not sensitive to sharpener design.}
    \label{tab:architecture}
\end{table}

\begin{figure}[H]
    \centering
    \includegraphics[width=\columnwidth]{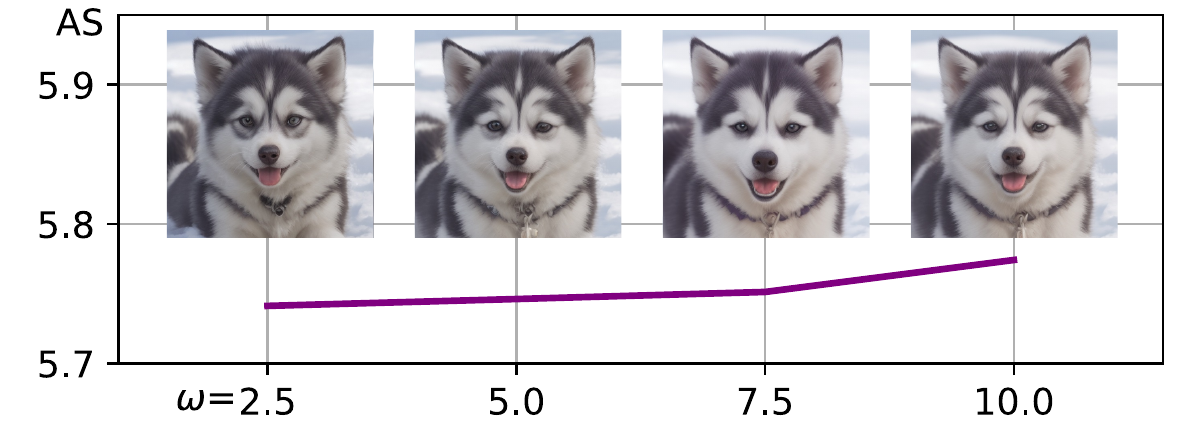}
    \caption{Sensitivity of guidance scale $\omega$ to Aesthetic Score (AS) during the training of \ourName. The performance of \ourName is not sensitive to the choice of $\omega$.
    }
    \label{fig:ablate_omega}
\end{figure}

\textbf{Sensitivity of guidance scale during training.}
The only hyperparameter in the training of \ourName is the guidance scale $\omega$. During training, we choose a guidance scale of 5, which is close to the recommended setting for mainstream text-to-image models. As shown in Figure \ref{fig:ablate_omega}, \ourName generates high-quality images across different guidance scales used in training, indicating that \ourName is not sensitive to the choice of guidance scale, avoiding the need for extensive hyperparameter tuning.

\begin{figure*}[t]
\begin{center}
\includegraphics[width=\textwidth]{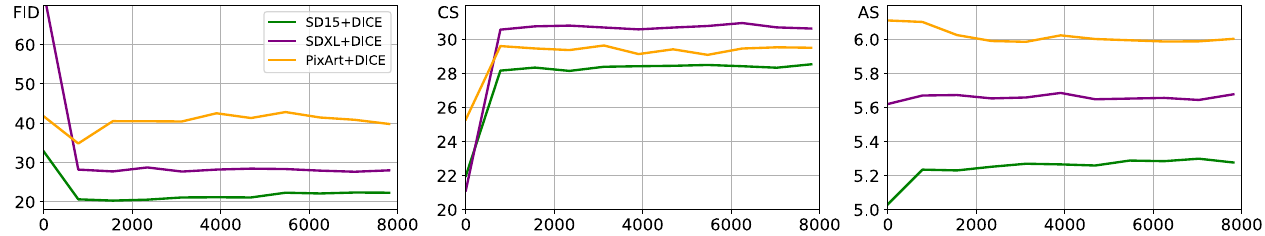}
\caption{Quantitative results on the training iterations of our method.}
\label{fig:ablation_iter}
\end{center}
\end{figure*}

\textbf{Training iterations.}
In Figure \ref{fig:ablation_iter}, we show the performance of our method with respect to FID, CS, and AS evolving with training iterations. Our method enjoys fast convergence.

\begin{figure*}[t]
\begin{center}
    \includegraphics[width=\textwidth]{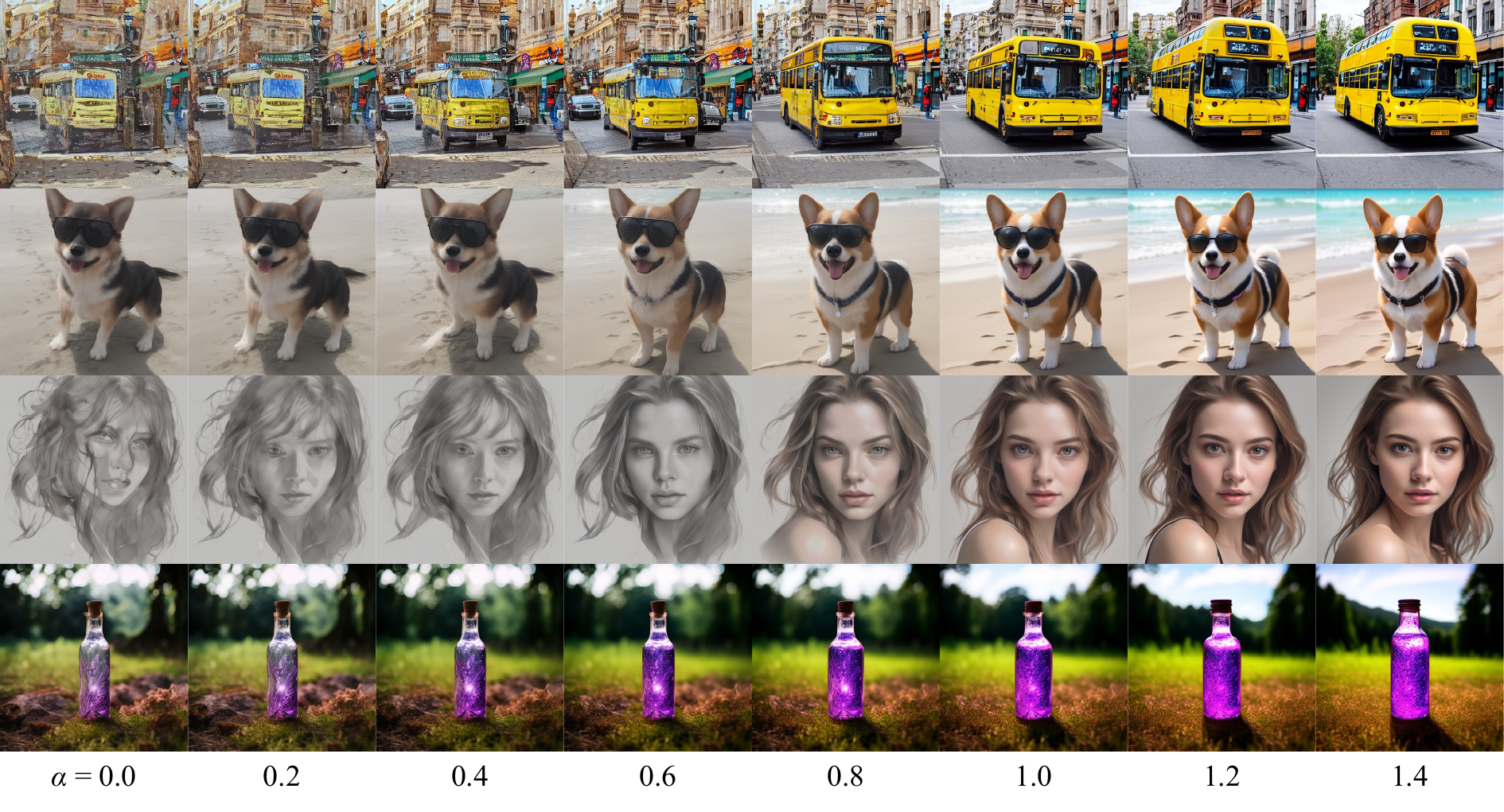}
    \caption{Qualitative results on sharpening strength $\alpha$. The sharpened text embeddings exhibit strong semantic consistency. 
    (1): SD15 with text prompt ``\textit{A yellow bus by the street, high quality, best details}''.
    (2): DreamShaper with text prompt ``\textit{A Corgi wearing sunglasses on the beach}''.
    (3): SDXL with text prompt ``\textit{A beautiful woman facing the camera, close, realistic}''.
    (4): PixArt-$\alpha$ with text prompt ``\textit{A glass bottle on the grass with purple galaxy inside}''.
    }
    \label{fig:ablation_alpha_qualitative}
\end{center}
\end{figure*}

\begin{figure*}[t]
\begin{center}
\includegraphics[width=\textwidth]{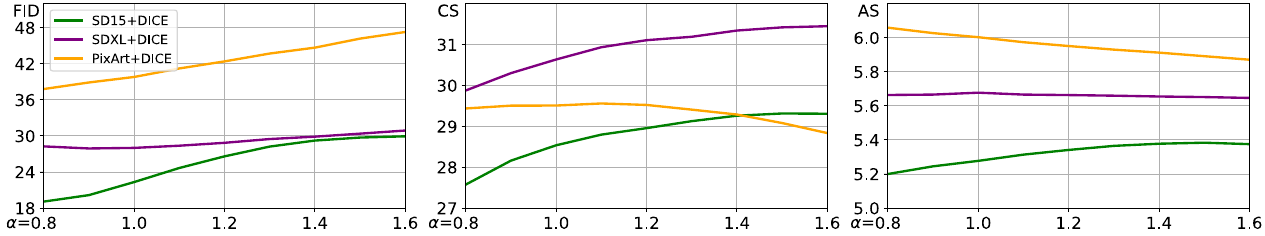}
\caption{Quantitative results on sharpening strength $\alpha$.}
\label{fig:ablation_alpha}
\end{center}
\end{figure*}

\textbf{Sharpening strength.} 
We conduct an ablation study on sharpening strength $\alpha$, which is fixed to 1 during training. The quantitative and qualitative results are shown in Figure \ref{fig:ablation_alpha} and Figure \ref{fig:ablation_alpha_qualitative}. As $\alpha$ increases in a certain range, the sample quality improves with richer detail and stronger contrast. 
The sharpener manages to find robust directions that are capable of enhancing the image quality while maintaining the semantic information.

\begin{figure*}[t]
    \includegraphics[width=\textwidth]{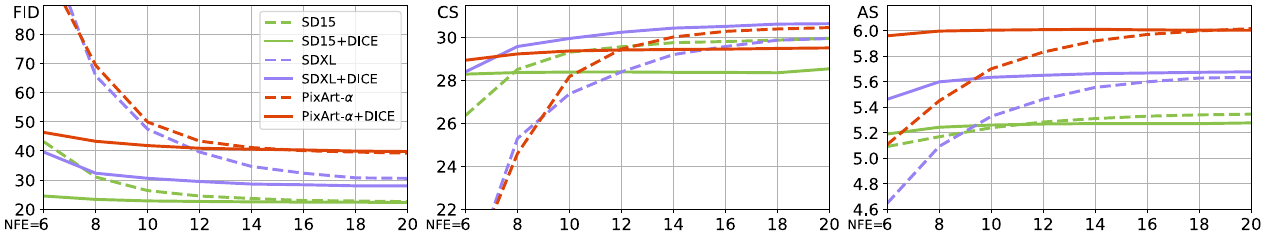}
    \caption{Comparison of FID (left), CLIP Score (middle), and Aesthetic Score (right) with respect to different numbers of function evaluations (NFE). \ourName converges faster than the guided sampling based on CFG. 
    }
    \label{fig:ablate_nfe}
\end{figure*}

\begin{figure*}[t]
\begin{center}
    \begin{subfigure}[t]{\textwidth}
        \centering
        \includegraphics[width=0.8\textwidth]{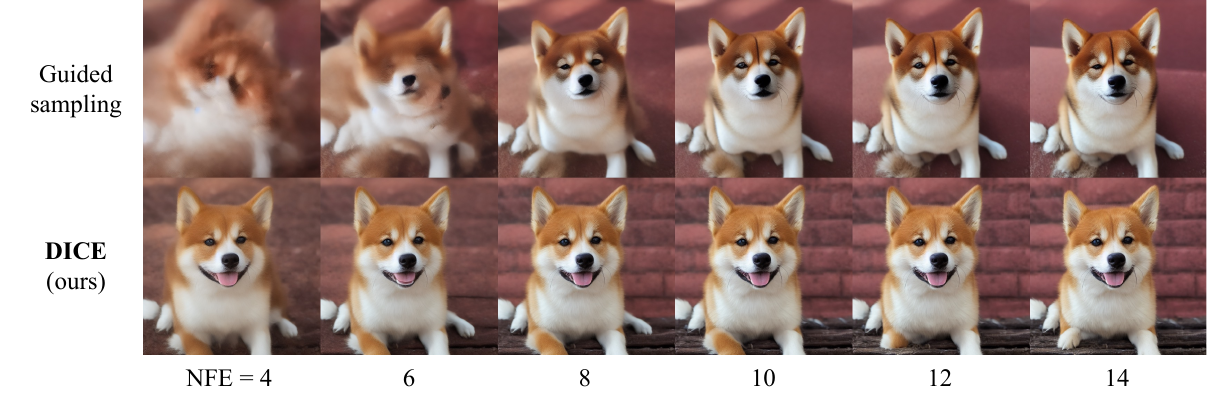}
        \caption{Model: Stable Diffusion v1.5~\cite{rombach2022ldm}. Text prompt: ``\textit{close-up photo of a cute smiling shiba inu}''.}
    \end{subfigure}
    \vskip 0.1in
    \begin{subfigure}[t]{\textwidth}
        \centering
        \includegraphics[width=0.8\textwidth]{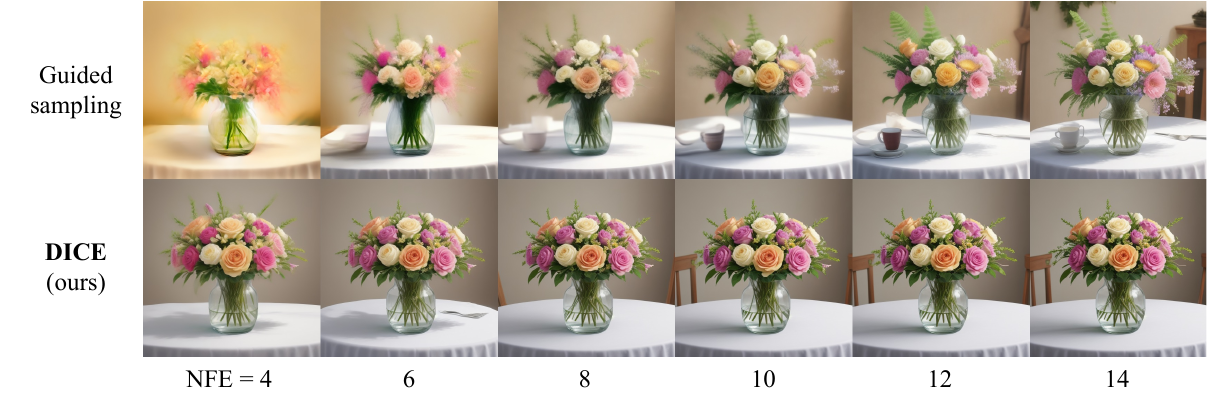}
        \caption{Model: DreamShaper. Text prompt: ``\textit{flowers on the table}''.}
    \end{subfigure}
    \vskip 0.1in
    \begin{subfigure}[t]{\textwidth}
        \centering
        \includegraphics[width=0.8\textwidth]{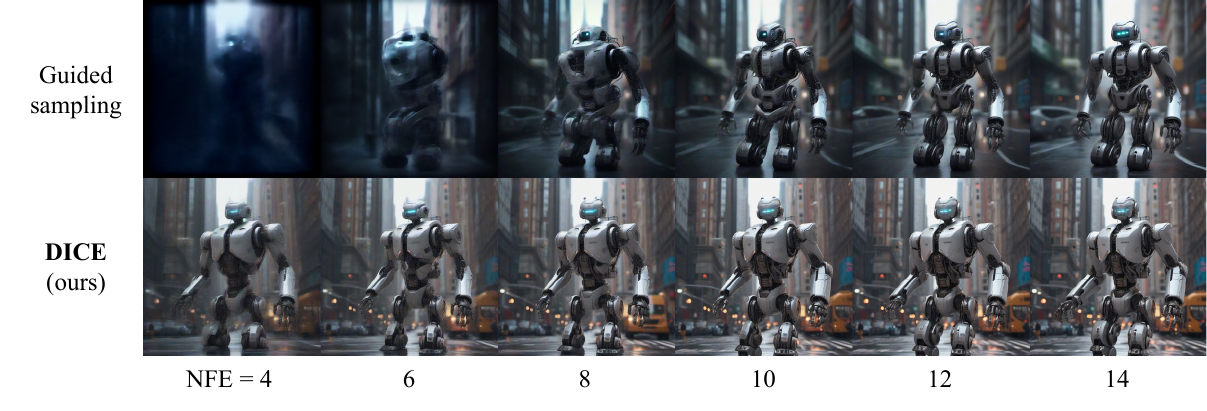}
        \caption{Model: Stable Diffusion XL~\cite{podell2024sdxl}. Text prompt: ``\textit{a robot in the city, perfect detail, 8k, best quality}''.}
    \end{subfigure}
    \vskip 0.1in
    \begin{subfigure}[t]{\textwidth}
        \centering
        \includegraphics[width=0.8\textwidth]{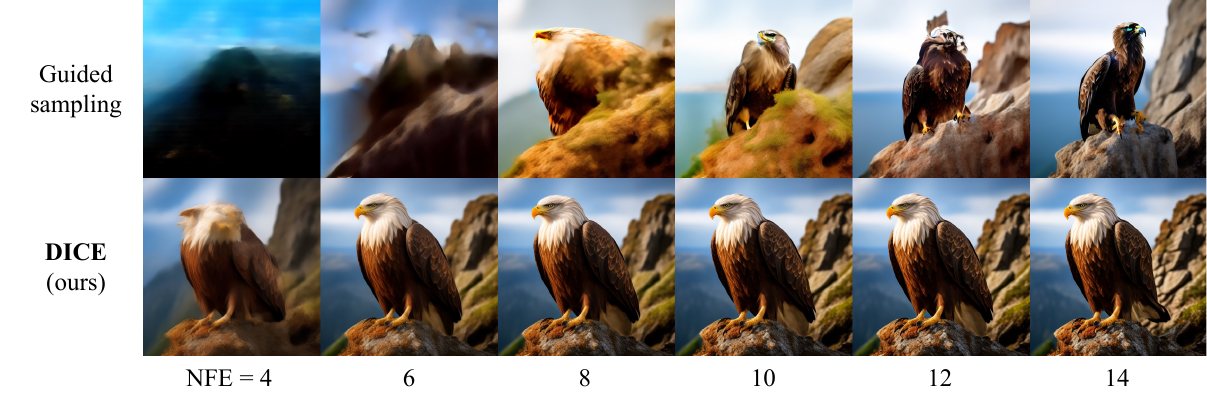}
        \caption{Model: Pixart-$\alpha$~\cite{chen2024pixart}. Text prompt: ``\textit{close-up photo of an eagle on the cliff}''.}
    \end{subfigure}
\caption{Qualitative results under different NFEs. A guidance scale of 5 is used for all guided sampling.}
\label{fig:ablate_nfe_qualitative}
\end{center}
\end{figure*}

\begin{figure*}[t]
\begin{center}
    \begin{subfigure}[t]{\textwidth}
        \centering
        \includegraphics[width=0.8\textwidth]{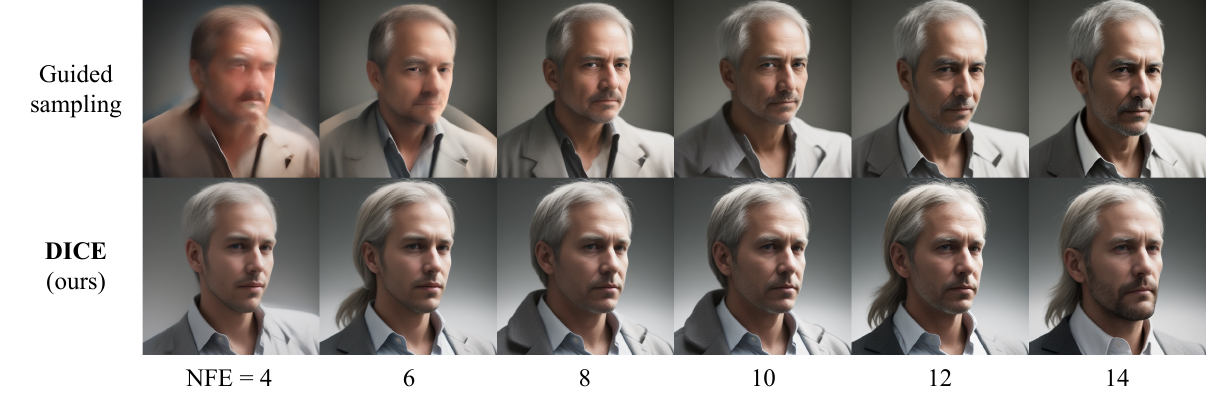}
        \caption{Model: AbsoluteReality. Text prompt: ``\textit{photo portrait of a man}''.}
    \end{subfigure}
    \vskip 0.1in
    \begin{subfigure}[t]{\textwidth}
        \centering
        \includegraphics[width=0.8\textwidth]{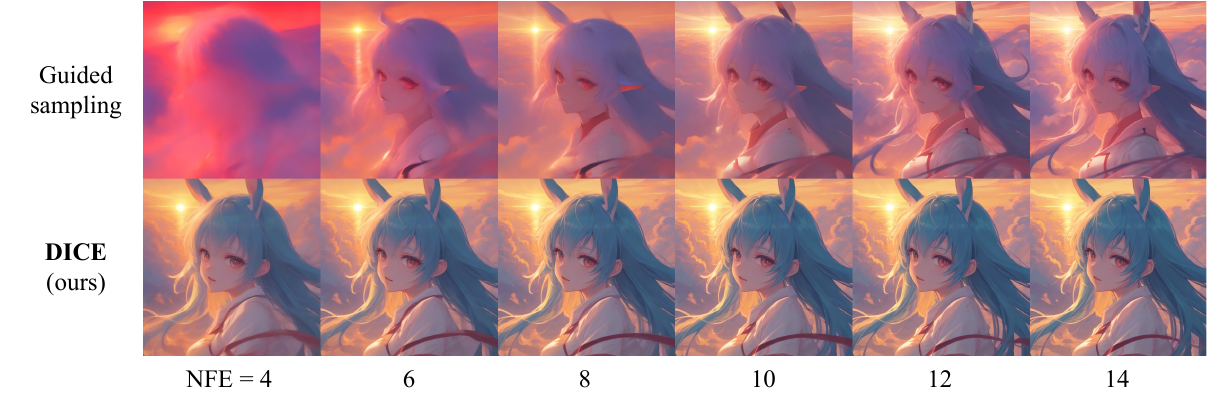}
        \caption{Model: Anime Pastel Dream. Text prompt: ``\textit{an anime character on the cloud, sunset, close-up}''.}
    \end{subfigure}
    \vskip 0.1in
    \begin{subfigure}[t]{\textwidth}
        \centering
        \includegraphics[width=0.8\textwidth]{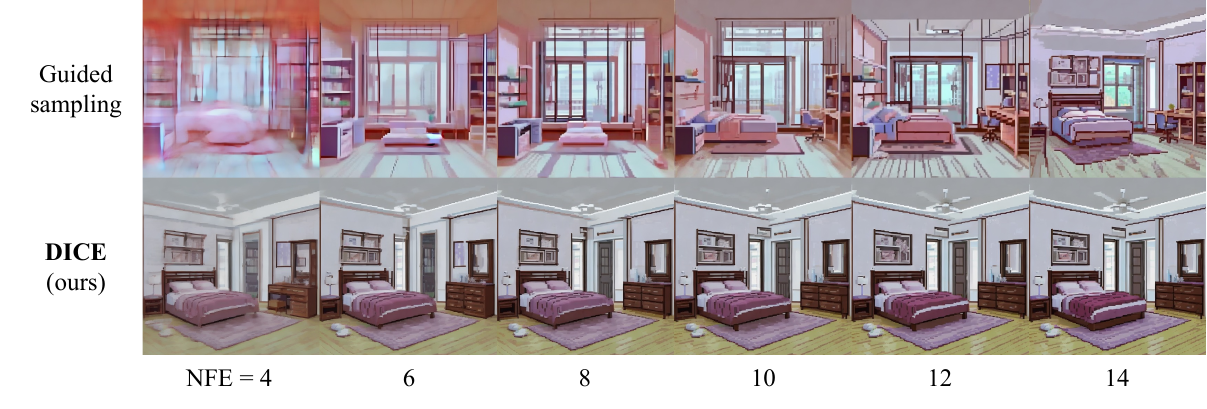}
        \caption{Model: DreamShaper PixelArt. Text prompt: ``\textit{a clean bedroom, pixel art}''.}
    \end{subfigure}
    \vskip 0.1in
    \begin{subfigure}[t]{\textwidth}
        \centering
        \includegraphics[width=0.8\textwidth]{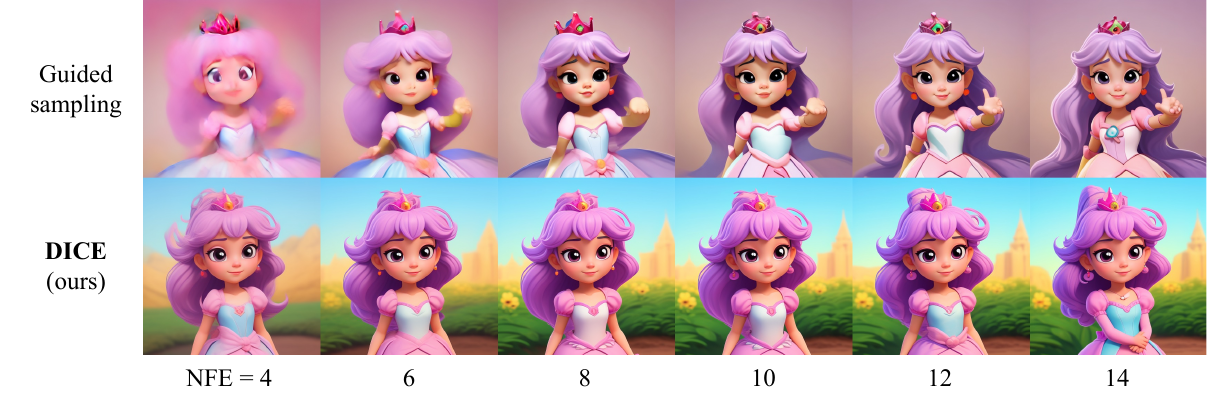}
        \caption{Model: 3D Animation Diffusion. Text prompt: ``\textit{a cute princess, cartoon, best quality}''.}
    \end{subfigure}
\caption{Qualitative results for other Stable Diffusion variants. Guidance scale of 5 is used for all guided sampling.}
\label{fig:ablate_nfe_qualitative_variants}
\end{center}
\end{figure*}

\textbf{NFE budgets.} 
In Figure \ref{fig:ablate_nfe}, we report the performance of \ourName and guided sampling evolving with the number of function evaluations (NFE). As CFG requires an additional model evaluation, the NFE for every single guided sampling step is two. 
\ourName exhibits superiority over guided sampling when operating under a low NFE budget.
Besides, the sample quality of \ourName exhibits an early convergence, which can be attributed to the smoother sampling trajectories~\cite{chen2024trajectory} generated by unguided sampling, as revealed in~\cite{zhou2024simple}. 
In Figure \ref{fig:ablate_nfe_qualitative} and Figure \ref{fig:ablate_nfe_qualitative_variants}, we provide qualitative results as a supplement to Figure \ref{fig:ablate_nfe} on all the text-to-image models involved in this paper, demonstrating the advantage of our method under low NFE budgets. 

\begin{figure*}[t]
\begin{center}
    \begin{subfigure}[t]{\textwidth}
        \includegraphics[width=\textwidth]{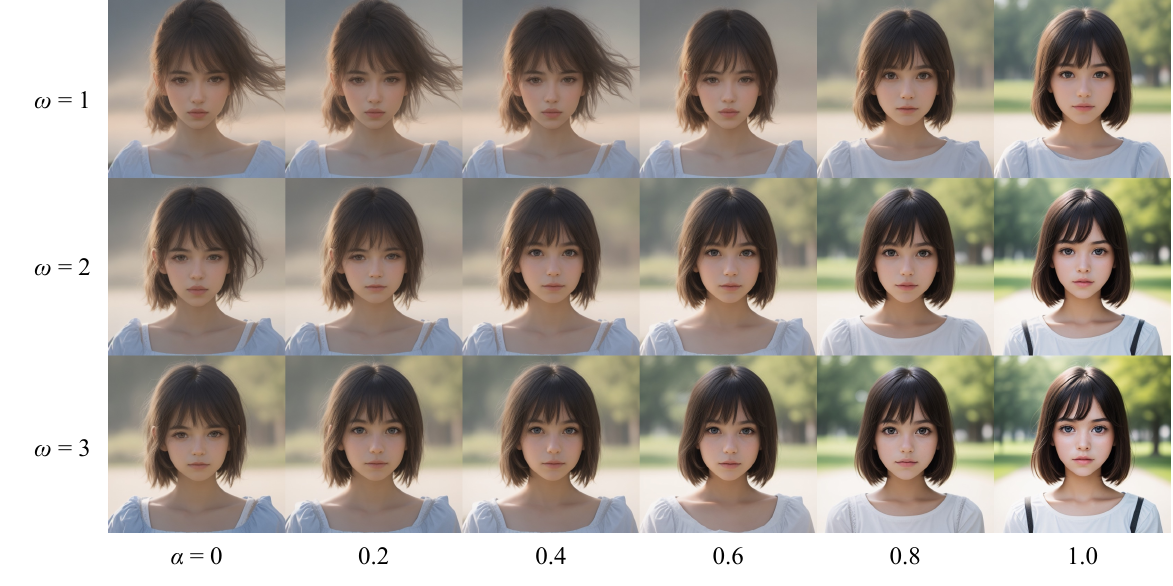}
        \caption{Model: DreamShaper. Text prompt: ``\textit{photo portrait of a girl}''.}
    \end{subfigure}
    \vskip 0.15in
    \begin{subfigure}[t]{\textwidth}
        \includegraphics[width=\textwidth]{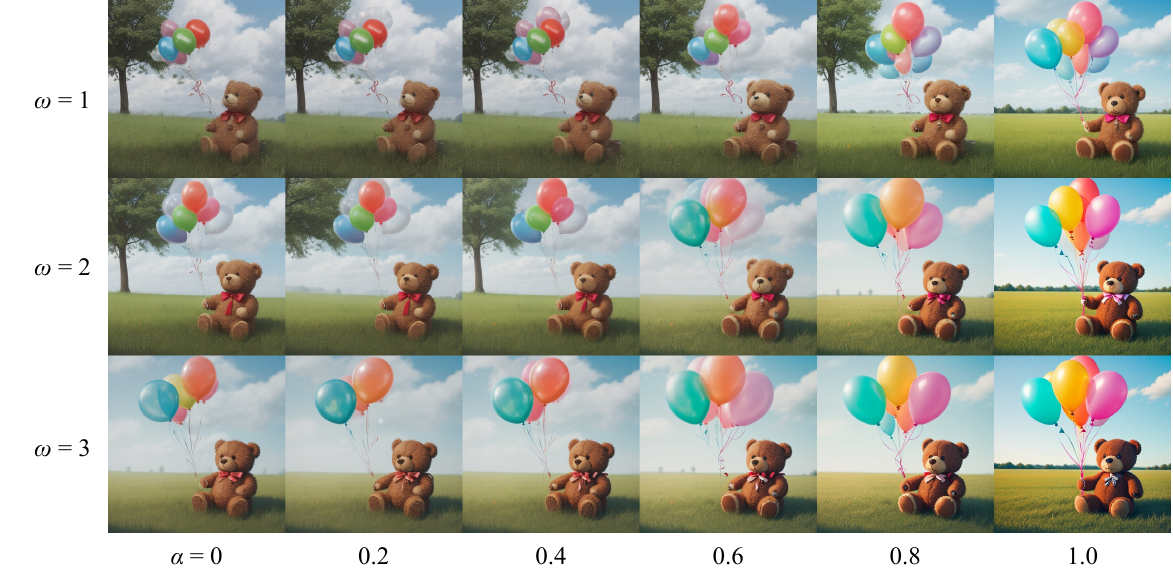}
        \caption{Model: DreamShaper. Text prompt: ``\textit{a teddy bear on the grass with balloons}''.}
    \end{subfigure}
\caption{Additional results on sharpening strength $\alpha$ and guidance scale $\omega$ during inference. Our sharpened text embeddings can also be combined with guided sampling by introducing guidance scale $\omega > 1$ back during inference to improve the overall image quality.}
\label{fig:ablate_cfg_alpha}
\end{center}
\end{figure*}

\textbf{Combined with guided sampling.} 
Our method is able to be combined with guided sampling by introducing a guidance scale $\omega > 1$ back during inference, as shown in Figure \ref{fig:ablate_cfg_alpha}. Combined with guided sampling, our method rapidly improves the image contrast as the guidance scale increases.

\begin{figure*}[t]
\begin{center}
    \begin{subfigure}[t]{\textwidth}
        \includegraphics[width=\textwidth]{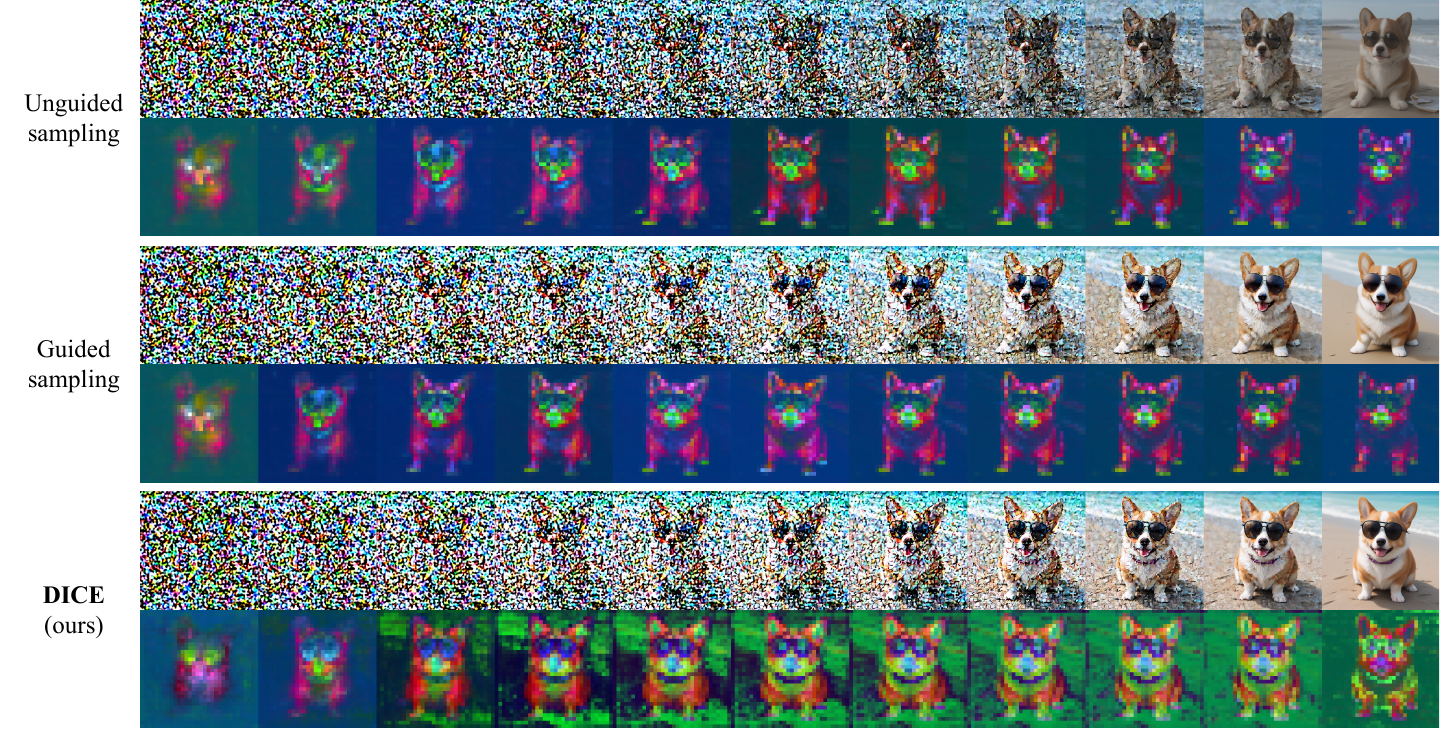}
        \caption{Model: DreamShaper~\cite{DreamShaper}. Text prompt: ``\textit{a Corgi wearing sunglasses on the beach}''.}
    \end{subfigure}
    \vskip 0.15in
    \begin{subfigure}[t]{\textwidth}
        \includegraphics[width=\textwidth]{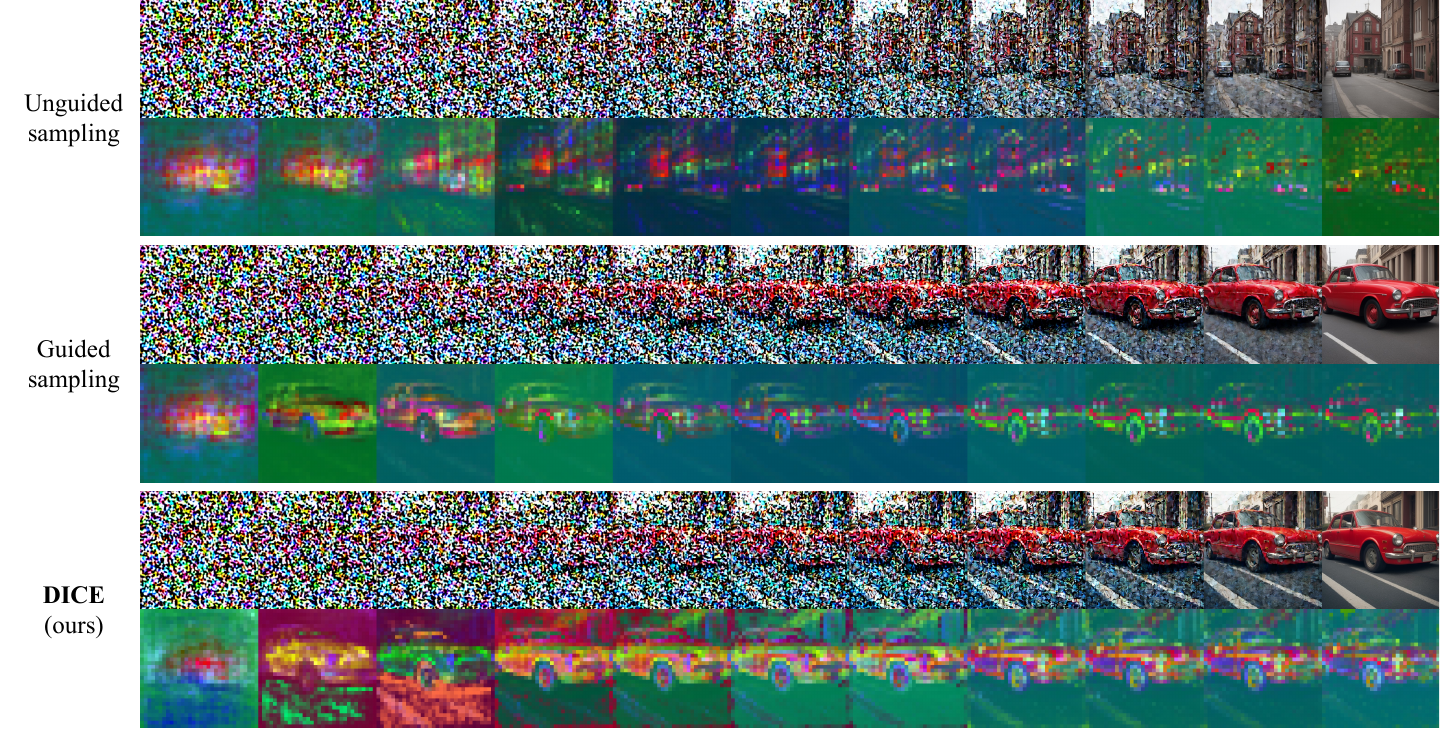}
        \caption{Model: DreamShaper~\cite{DreamShaper}. Text prompt: ``\textit{a red car by the street}''.}
    \end{subfigure}
\caption{Qualitative results on sampling trajectory and cross-attention maps from $t=T$ to $t=0$. Images are generated by 10-step DPM-Solver++~\cite{lu2022dpmpp} with the same random seed in each subfigure.}
\label{fig:traj}
\end{center}
\end{figure*}

\textbf{Cross-attention maps.} 
We visualize the cross-attention maps in Figure \ref{fig:traj} to show how sharpened text embeddings affect sampling dynamics. Specifically, we apply principal component analysis to the extracted feature maps from the Stable Diffusion U-Net decoder~\cite{rombach2022ldm} and use the top three principal components to compose RGB images for visualization. While the main concepts (i.e., Corgi and sunglasses) are activated in all methods, the sharpened embedding further activates image details in the background. This also aligns with our conclusion of the mechanism of \ourName drawn above.

